# Automatic Discrimination of Color Retinal Images using the Bag of Words Approach

*Ibrahim Sadek*

**Supervisors:**

Professor. Fabrice Meriaudeau
Associate Professor. Désiré Sidibé

Centre Universitaire Condorcet
Université de Bourgogne, Le Creusot, France

A Thesis Submitted for the Degree of
MSc Erasmus Mundus in Vision and Robotics (VIBOT)

· 2014 ·

**Abstract**

Diabetic retinopathy (DR) and age related macular degeneration (ARMD) are among the major causes of visual impairment worldwide. DR is mainly characterized by red spots, namely microaneurysms and bright lesions, specifically exudates whereas ARMD is mainly identified by tiny yellow or white deposits called drusen. Since exudates might be the only manifestation of the early diabetic retinopathy, there is an increase demand for automatic retinopathy diagnosis. Exudates and drusen may share similar appearances, thus discriminating between them is of interest to enhance screening performance. In this research, we investigative the role of bag of words approach in the automatic diagnosis of retinopathy diabetes. We proposed to use a single based and multiple based methods for the construction of the visual dictionary by combining the histogram of word occurrences from each dictionary and building a single histogram. The introduced approach is evaluated for automatic diagnosis of normal and abnormal color fundus images with bright lesions. This approach has been implemented on 430 fundus images, including six publicly available datasets, in addition to one local dataset. The mean accuracies reported are 97.2% and 99.77% for single based and multiple based dictionaries respectively.

*Imagination is more important than knowledge....*

Albert Einstein

# Contents

# List of Figures

# List of Tables

# Acknowledgments

First of all, I would like to express my deep gratitude to my supervisors and all the people that helped in this project: Prof. Fabrice Meriaudeau, Dr. Désiré Sidibé. Thanks, for their support, guidance and patience for the success of this work. Special thanks should be given to my Erasmus Mundus mobility grant program (Erasmus Mundus in Computer and Robotics - VIBOT). I would like to thank the program coordinator Dr. David Fofi for his guidance and support through the entire period of my master. And last, but not least important, very special thanks go to my beloved wife Lamees, parents, who supported and encouraged me during stressful moments.

# Chapter 1

# Introduction

According to the world health organization (WHO) diabetes mellitus (DM) is a lifelong disorder which takes place either when the pancreas doesn't produce sufficient insulin (type 1 diabetes) or when the body cannot effectively benefit the insulin it produces (type 2 diabetes). Insulin is a hormone produced in the pancreas by beta cells that regulates the level of blood sugar. Hyperglycemia, or increased blood sugar level causes serious damage to body's system, including diabetic retinopathy. The most important reasons of diabetes are increasing age, overweight, and sedentary lifestyle.

During the first two decades of disease, approximately all patients with type 1 diabetes and more than 60% of patients with type 2 diabetes have retinopathy [1]. The prevalence of diabetes is estimated to increase from 2.8% to 4.4% in the time span of $2000 - 2030$. The total number of people is projected to increase from 171 million in 2000 to 360 million in 2030 [2]. Diabetic patients can prevent severe visual loss by attending regular diabetic eye screening programs and receiving optimal treatments [3].

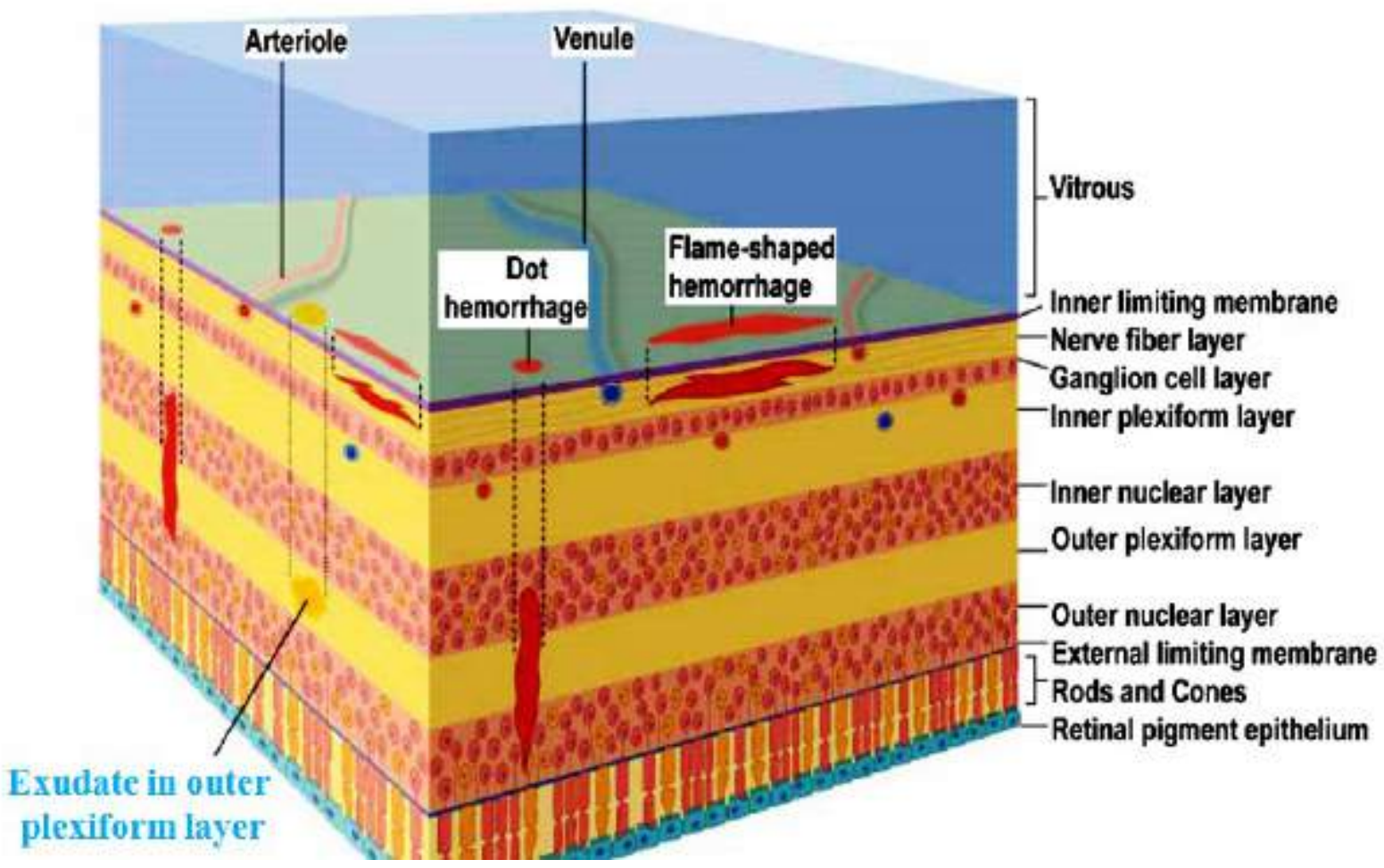

Figure 1.1: Exudate example in the outer layer of the retina [4].

Diabetic retinopathy (DR) and age related macular degeneration (ARMD) are among the leading causes of visual impairment worldwide. DR occurs most frequently in adult aged $(20-74)$ years, and it is characterized by the presence of red lesions (microaneurysms) and bright lesions (exudates) which appear as small white or yellowish white deposits with sharp margins and variable shapes located in the outer layer of the retina, their detection is essential for diabetic retinopathy screening systems. ARMD usually affects people over 50 years of age. It is caused by a damage to the macula, the small sensitive area of the retina that gives central vision (seeing fine details and colors), and categorized by drusen, tiny yellow or white deposits in a retina layer called Bruch's membrane. The Severity of ARMD can be categorized into three classes: early, intermediate, and advanced as discussed in Section 2.5. Two examples of exudate and drusen in a human retina are shown in Fig. 1.1 and Fig. 1.2 respectively.

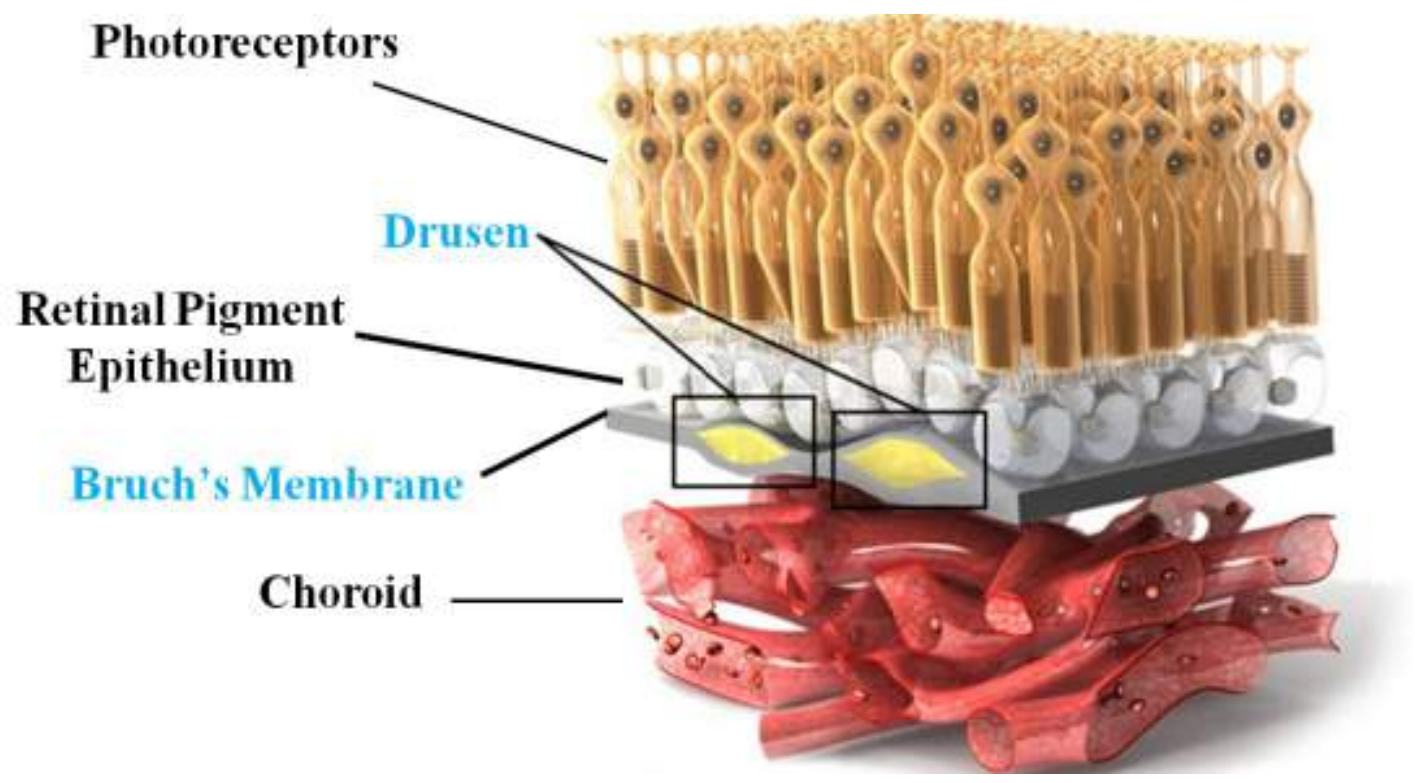

Figure 1.2: Drusen example within Bruch's membrane [5].

In some patients bright lesions such as retinal exudates can be the only manifestations of early diabetic retinopathy. Thus, computer aided detection (CAD) systems have been proposed in order to detect exudates. However, these bright lesions must be identified from drusen because they share common characteristics [6]. This represents a challenge for readers or CAD based screening systems designed for DR diagnosis. Consequently, developing a CAD system for reading and analyzing retinal images decreases observational unintentional failure and the false negative rates of ophthalmologists interpreting these images.

## 1.1 Aims and objectives

The aim of this research work is to design a system that will be able to identify normal, drusen, and exudates in color retinal fundus images using the bag of words approach (BOW), since there is a few approaches in the literature designed for this purpose. The thesis overview is explained as follows:

- Image preprocessing in order to correct uneven illumination and enhance contrast between lesions in the retina and other structures.
- Extracting different features from the training set.
- Building a visual vocabulary and finding a histogram of word frequencies in the vocabulary for constructing the BOW feature.

- Training a classier and finding the best parameters by carrying out a class classification with 10 fold cross validation.
- Measuring the performance of the system.

## 1.2 Thesis Structure

This thesis is structured as discussed below:

1. **Chapter 1** briefly describes the problem statement and the objective of this project.
2. **Chapter 2** briefly describes the human eye anatomy and the overall arrangement of retinal layers, different retinal imaging techniques, clinical lesions of the retina i.e. cotton wool spots, hard exudates, drusen, microaneurysms. Besides, diabetes mellitus, and diabetic retinopathy.
3. **Chapter 3** presents a state of the art review, different methods used to differentiate between drusen and exudates.
4. **Chapter 4** the stages involved in DR diagnosis are discussed. These stages include preprocessing, lesions classification using the BOW approach.
5. **Chapter 5** provides results and discussion.
6. **Chapter 6** gives conclusions and recommendations for the future work.

## 1.3 Project Planning

The duration of the master thesis is 4 months, beginning at 3rd of February 2014. The master planning is organized as follows:

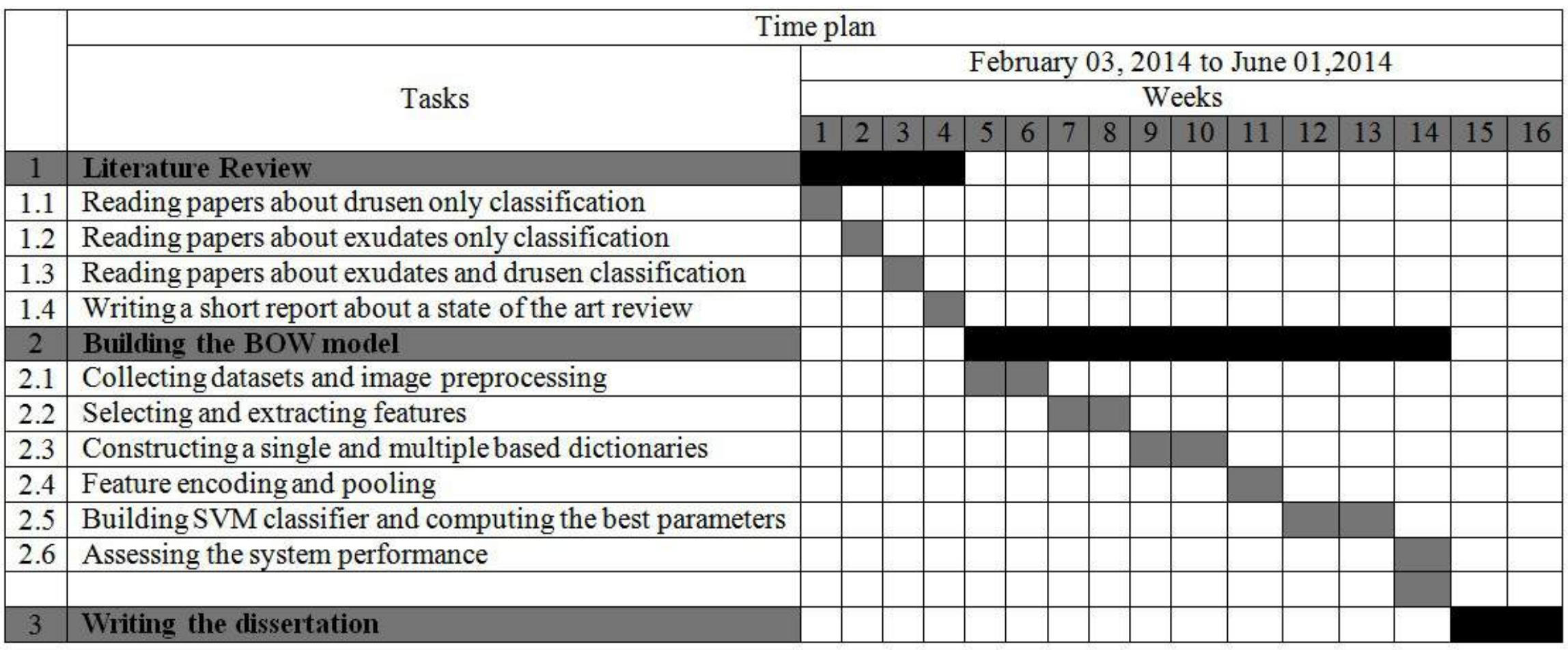

Figure 1.3: Master thesis planning.

# Chapter 2

# Background

This chapter is organized such as. Section 2.1 briefly describes the human eye anatomy and the overall arrangement of retinal layers. Section 2.2 introduces different retinal imaging techniques. Section 2.3 discusses clinical lesions of the retina i.e. cotton wool spots, hard exudates, drusen, microaneurysms. Section 2.4 describes diabetes mellitus. Section 2.5 presents associated diabetic disorders and its manifestations.

## 2.1 Eye Anatomy

Human eye is responsible for the detection of visible light, the part of the electromagnetic spectrum with wavelengths ranging approximately from 400 to 700 nm as shown in Fig. 2.1. The color of visible light depends on its wavelength. For example, violet has a wavelength of 400 nm, and red has a wavelength of 700 nm [7].

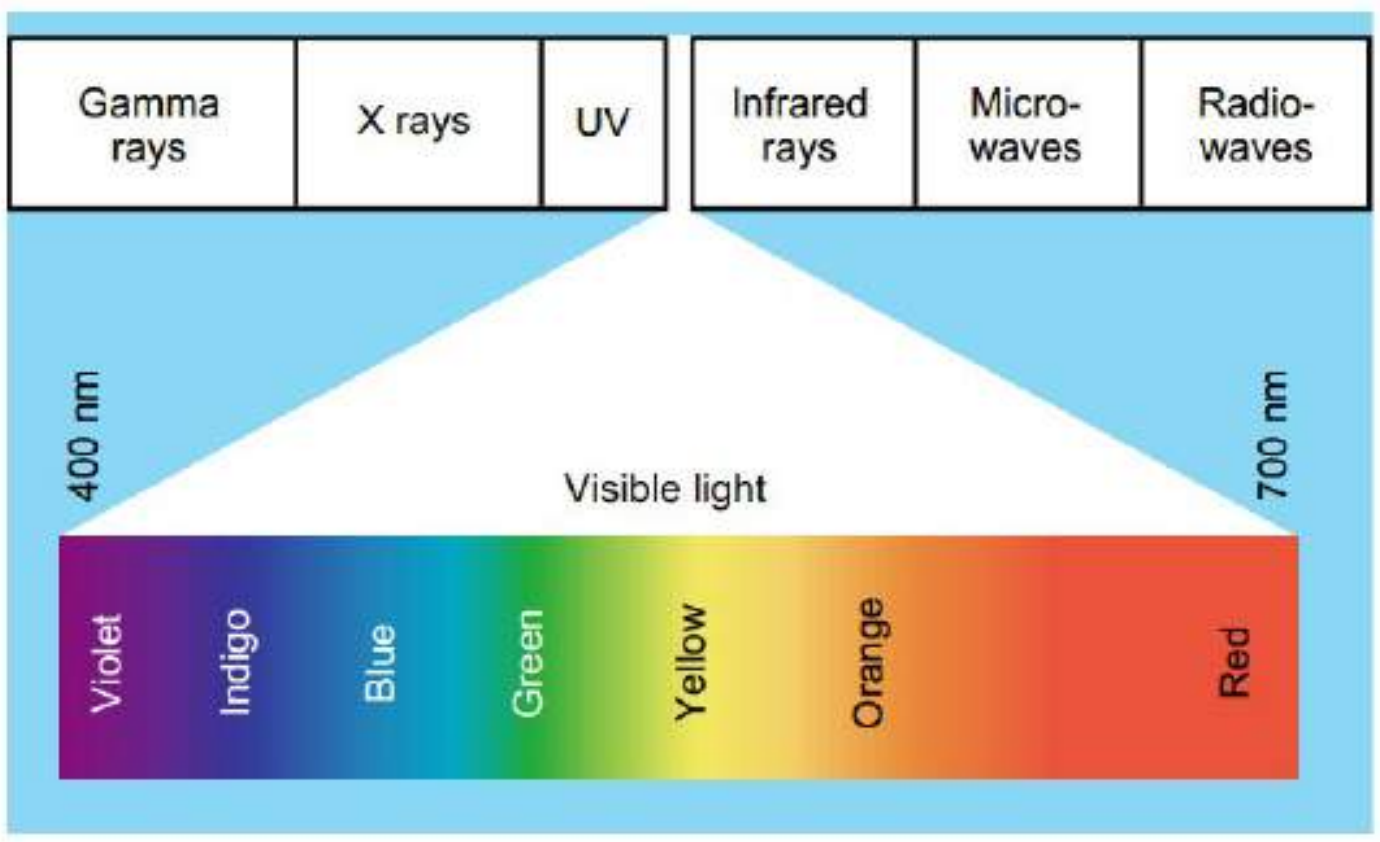

Figure 2.1: Electromagnetic spectrum [7].

Apparently, most people would agree that vision is much more valuable than other sense organs present in our body, because we usually use our eyes in almost all daily activities we perform. There exist many similarities between the human eye and a camera. For instance, in a camera, a film is used to record the image; in the eye the image is focused on the retina; also as the shutter in a camera, the pupil at the center of the iris control the amount of light that gets through the lens [8].

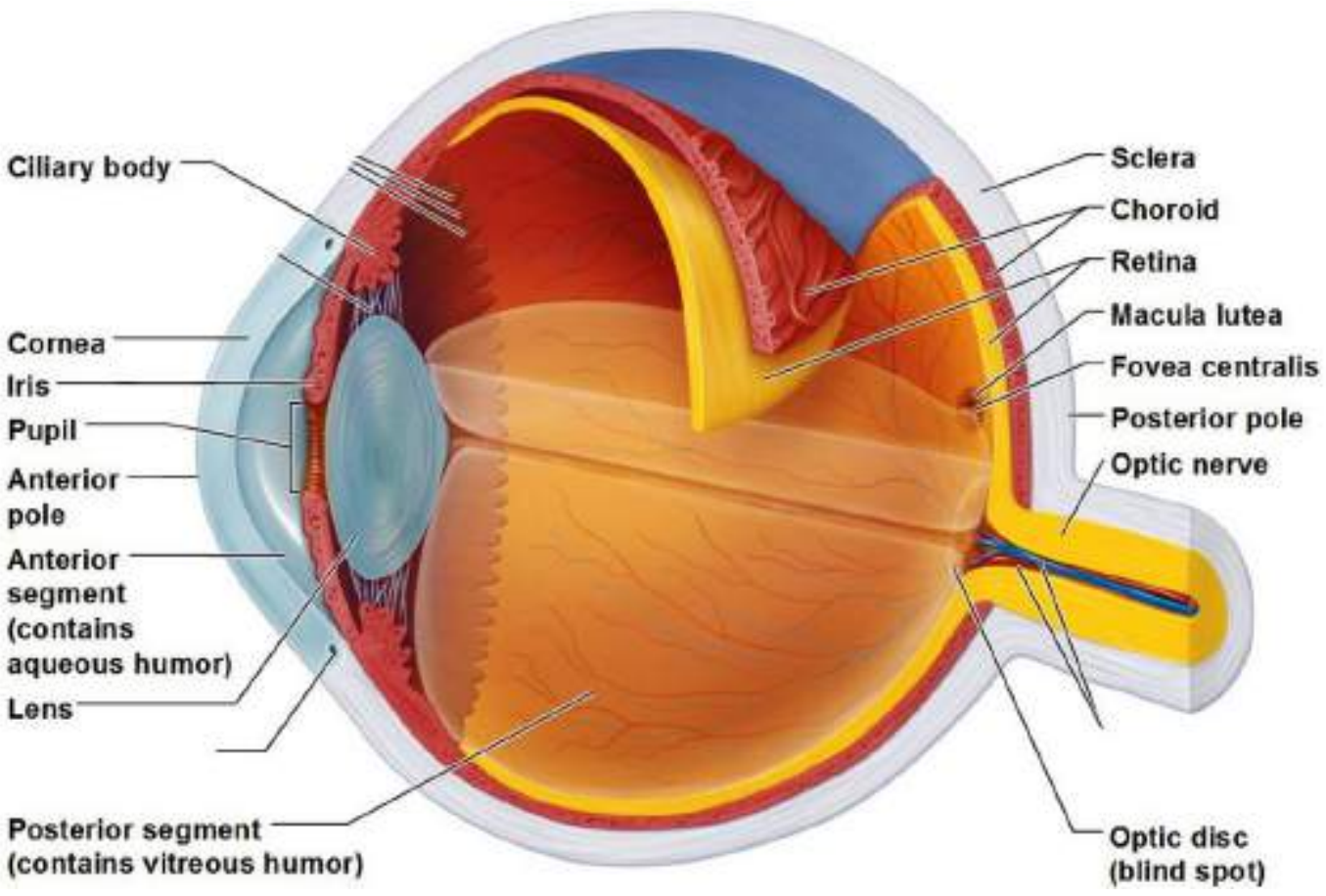

Figure 2.2: Cross section of a human eye anatomy.

Fig. 2.2 presents a cross section of a human eye anatomy. When light enters the eye, it first comes in contact with the cornea. The cornea filters and refracts the light, allowing the image to converge inside the eye to its way to the iris and pupil. The iris will constrict or dilate (widen) in order to adjust the pupil size, as a result regulating the amount of light that can enter the eye. The lens can change its shape with the help of auxiliary muscles and bring objects into focus, also it slightly improves the already refined image from the cornea and projects it onto the retina [9]. The structures of retina are more related to our research work among the several ocular structures, hence it will be elaborated in depth as in Section 2.1.1.

### 2.1.1 Retina

The retina is the light sensitive tissue that covers the interior surface of the eye. The cornea and lens focus light rays on the retina. Then, the retina converts the light received into electrical impulses that are sent via the optic nerve to the brain which interprets them as images. The cornea and lens in the eye behave like the camera lens, while the retina is analogous to the film. If the images are not focused properly, the film or retina receives blurry images. The retina consists of two main photoreceptors; the rods and the cones [10]. Table 2.1 shows the main differences between the rods and cones.

| Rods | Cones |
|---|---|
| Require a very low level of light to generate signals. | Require a higher level of light to generate signals. |
| Approximately 125 millions photoreceptors. | Approximately 6 millions photoreceptors. |
| Specialized for low light vision. | Mediate daylight and color vision. |
| Distributed in the periphery of the retina. | Concentrated in the fovea, which lies in the center of the macula. |

Table 2.1: Differences between rods and cones in a human retina.

The retina can be divided into many distinguishable layers as shown in Fig. 2.3.

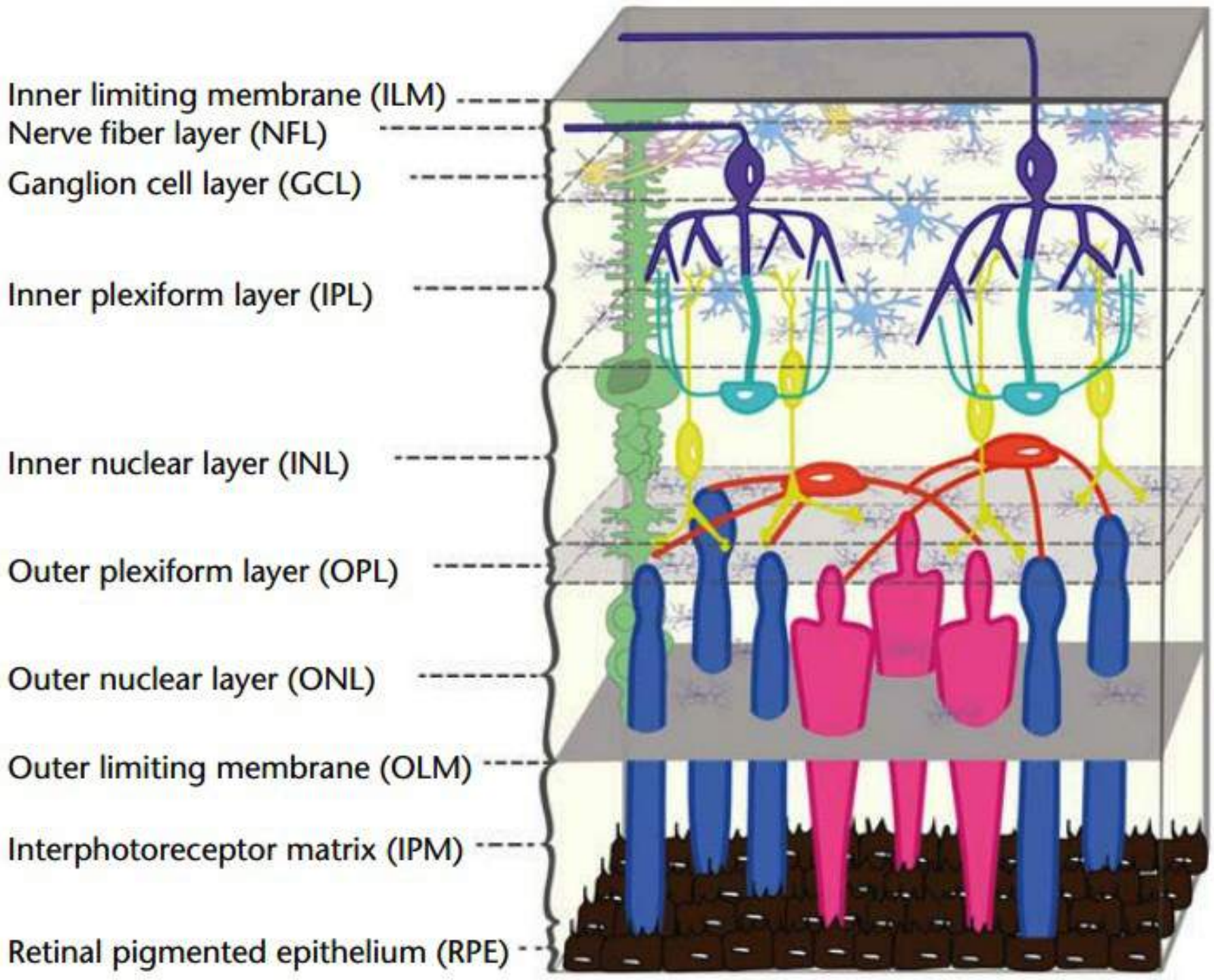

Figure 2.3: Schematic view of retina layers organization [11].

The different layers of retina are organized as follows [12]:

1. **The Inner Limiting Membrane (ILM)** is the boundary between vitreous body and the retina.
2. **Ganglion Cell Layer (GCL)** encompasses the cell bodies and axons of the ganglion cells.
3. **Inner Plexiform Layer (IPL)** comprises the synapses between bipolar, amacrine, and ganglion cells.
4. **Inner Nuclear Layer (INL)** involves bipolar cell, horizontal and amacrine cell bodies.
5. **Outer Plexiform Layer (OPL)** consists of bipolar cells, horizontal and receptor synapses.
6. **Outer Nuclear Layer (ONL)** contains the nuclei of photoreceptor.
7. **Outer Limiting Membrane (OLM)** comes into contact with the base of the inner segments of photoreceptor.
8. **Photoreceptor Layer** contains the inner and outer segments of photoreceptors.
9. **The Pigment Epithelium Layer** is the outermost layer of the retina be composed of pigmented cuboidal cells that contain Melanin. Melanin is the black pigment which absorbs any excess light that is not captured by the retina and prevents it from reflection back to the retina. Thus, protects the photo-receptors from damaging level of light. The pigment epithelium cells provide nutrition such as glucose, and essential ions to photoreceptors.

The ganglion cells are a type of neuron receiving visual information from photoreceptors through two intermediate neurons (horizontal, and amacrine). The horizontal cells which are interconnecting neurons, help integrate and regulate the input from multiple photoreceptor cells.

## 2.2 Retinal Imaging Techniques

Retinal images are acquired by highly specialized camera called fundus camera. As a matter of fact, retinal images play an important role in diagnosing several eye diseases such as diabetic retinopathy, glaucoma, and age related macular degeneration. For many years, ophthalmoscopy, fundus photography, fluorescein angiography, and diagnostic ultrasound were considered the only retinal imaging modalities. Recently, different techniques have been introduced including indocyanine angiography, optical coherence tomography (OCT), scanning laser ophthalmoscopy (SLO), and infrared imaging. Additionally, digital technology permits easy data processing and storage as well as worldwide images exchange [13].

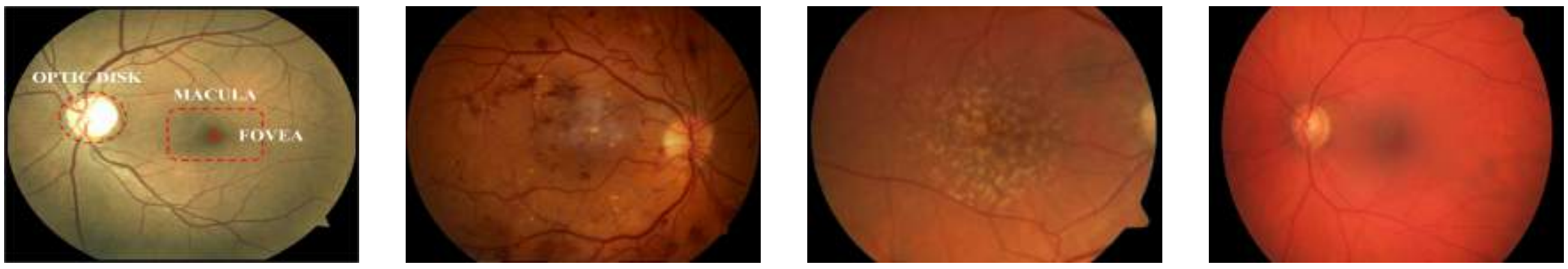

(a) (b) (c) (d)

Figure 2.4: Examples of retinal images; (a) healthy, (b) diabetic retinopathy, (c) age related macular degeneration, (d) glaucoma retinal images.

Typically, ophthalmoscopy (fundoscopy) is used to determine the health of the retina and vitreous humor. There are two types of ophthalmoscopes: direct and indirect. The first, is an instrument about the size of a small flash light with several lenses and a magnification up to 15 times. The second, is provided with a light attached to a head which can provide a wider view of the inside of the eye.

Fundus Photography is frequently used to create a picture of the interior surface of the eye, involving the retina, optic disc, and macula through the dilated pupil of the patient in order to enhance the quality of the image. It helps transfer images via networks for remote viewing by specialists or trained medical professionals. In addition, it allows for large scale screening programs which is a key in preventing diabetic retinopathy and induced vision deterioration that can lead to blindness. A fundus camera is a low power microscope with an attached camera which provides an upright magnified view of the fundus. An angle of $30^\circ$, considered the typical angle of view, creates $2.5x$ magnification. Wide angle fundus cameras capture images between $45^\circ$ to $140^\circ$ and provide less magnification. Whereas, a narrower angle fundus cameras have $20^\circ$ or less angle of view [14]. Fig. 2.4 shows different examples of retinal fundus images.

According to [15] fundus imaging can be defined as any process involving 2D representation of the 3D retinal structures projected onto the image plane and obtained using reflected light, such that image intensities represent the amount of reflected quantity of light. Furthermore, techniques/modalities belonging to fundus imaging are grouped together as discussed in Table 2.2.

| Name | Description |
|---|---|
| Fundus Photography | At a specific wavelength, image intensities correspond to the amount of reflected light. This includes red free, where the red color is filtered out from the imaging light, enhancing contrast of vessels and other structures. |
| Color Fundus Photography | The retina is fully examined in color by means of white light illumination. Subsequently, image intensities represent the amount of reflected red, green, and blue wavebands. |
| Stereo Fundus Photography | Allows depth estimation by exploiting two or more different view angles. |
| Hyperspectral Imaging | Produces multiple specific wavelength bands. Potential diseases may be indicated by monitoring oxygen consumption in the retina. |
| Scanning Laser Ophthalmoscopy (SLO) | Provides sharp retinal images by a single wavelength laser light obtained sequentially in time which can reach the surface of the retina and records its surface details. |
| Adaptive Optics Scanning Laser Ophthalmoscopy (AOSLO) | Optical light is corrected by modeling the deviation in its wavefront. |
| Fluorescein Angiography | Image intensities constitute the amounts of emitted photons from photosensitive materials i.e. fluorescein or indocyanine green fluorophore injected into the patient blood stream. |

Table 2.2: Different fundus imaging modalities [15].

Ophthalmic ultrasound, becomes increasingly important when doctors cannot see the retina due to bleeding inside the eye, severe cataract, or corneal scarring. This technique is similar to typical ultrasound used to scan different organs of human body, sound waves are sent from a probe placed on the eye to provide sonar images of the inside of the eye [16].

Essentially, optical coherence tomography (OCT) is based on the fundamentals of low coherence interferometry, in order to measure the echo time delay of the back-scattered light in the sample in the same manner as A-Scan ultrasound acquisition. A temporally and spatially, low coherent beam of light (generated from laser diode) split into two disjoint paths one is directed into the sample, while the other is directed into a reference mirror (at known, but variable distance from the source). Light from both beams are then reflected and overlap withing a fiber-optic interferometer [17]. Fig. 2.5 shows an example of an OCT image of a normal human macula.

## 2.3 Clinical Lesions of the Retina

In this section, two groups of retinal diseases will be discussed: yellow white spots, and red spots. The former, includes cotton wool spots, hard exudates, and drusen. The latter, consists of microaneurysms, and hemorrhages [18, 19].

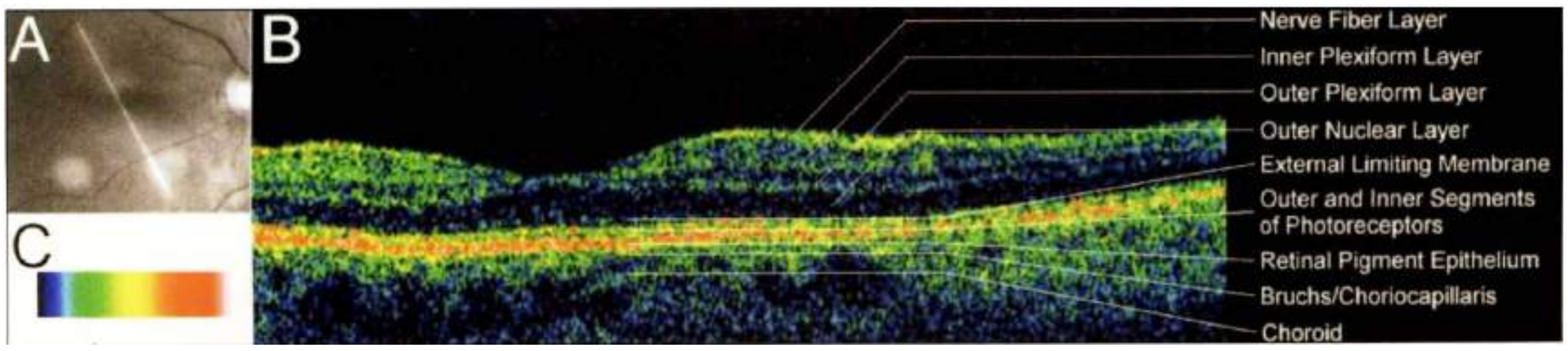

Figure 2.5: An OCT scan of a normal human macula; (a) a frame captured at the start of the scan assists in location, (b) an OCT of the specified scan location, (c) color map representing the log function of the reflectivity encountered by the probe beam. [17].

### 2.3.1 Soft Exudates or Cotton Wool Spots (CWS)

Soft Exudates or Cotton Wool Spots appear as white, feathery spots with fuzzy borders. CWS physically correspond to small retinal closures (infarcts) and swellings of the retinal nerve fiber layer because of microvascular diseases. They are located in the superficial (inner) retina, so they may obscure nearby vessels. Fig. 2.6(c) shows an example of such lesions.

### 2.3.2 Hard Exudates (HE)

Hard Exudates are lipoprotein and other kinds of protein originating from leaking microaneurysms. HE appear as small white or yellowish white deposits with sharp edges, irregular shape, and variable size as presented in Fig. 2.6(b). The physical locations of these lesions are deeper in the retina than cotton wool spots.

### 2.3.3 Drusen

Drusen are variable sized yellowish lipoproteinaceous deposits that form between the retinal pigmented epithelium (RPE), and Bruch's membrane. Usually , drusen alone do not contribute to vision loss. However, an increase in the size or number of such lesions are the earliest signs of age related macular degeneration (ARMD). Fig. 2.6(f) shows an example.

### 2.3.4 Microaneurysms (MAs)

Microaneurysms are among the earliest noticeable indication of retinal damage. MAs appear as small, round and dark red dots on the retinal surface that have sharp margins. By definition, their sizes are less the main optic veins as they cross the optic disc. They are the physical dilation (weakening) of the capillary walls which stimulate them to leakages. Fig. 2.6(e) shows an example of two MAs lesions.

### 2.3.5 Hemorrhages (HEM)

Hemorrhages happen due to leakage of weak capillaries i.e. bleeding under the conjunctiva which is the outermost protective layer of the eyeball. Generally, they have several shapes such as dot, blot, and flame. HEM are described as red spots with uneven, or indistinct edges and coloring. Their sizes are greater than MAs sizes as shown in Fig. 2.6(b).

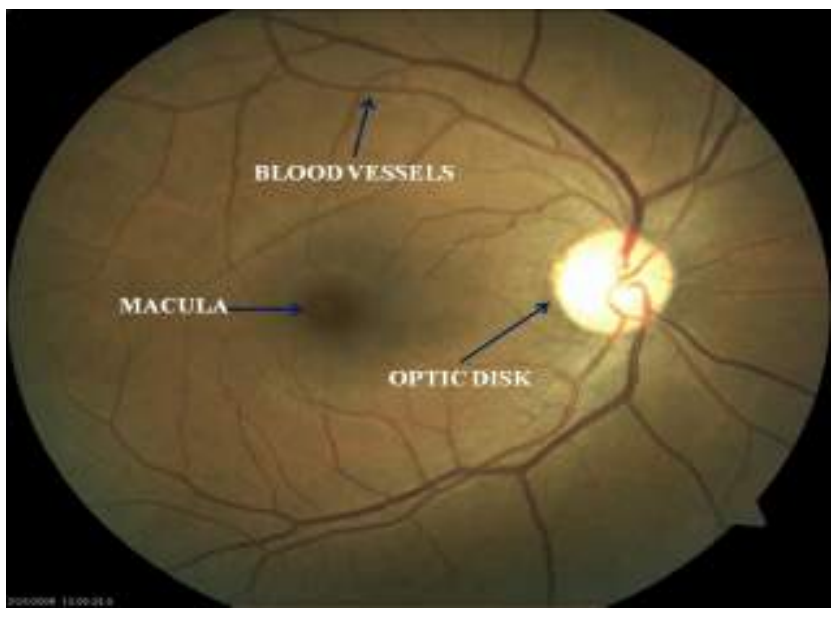

(a)

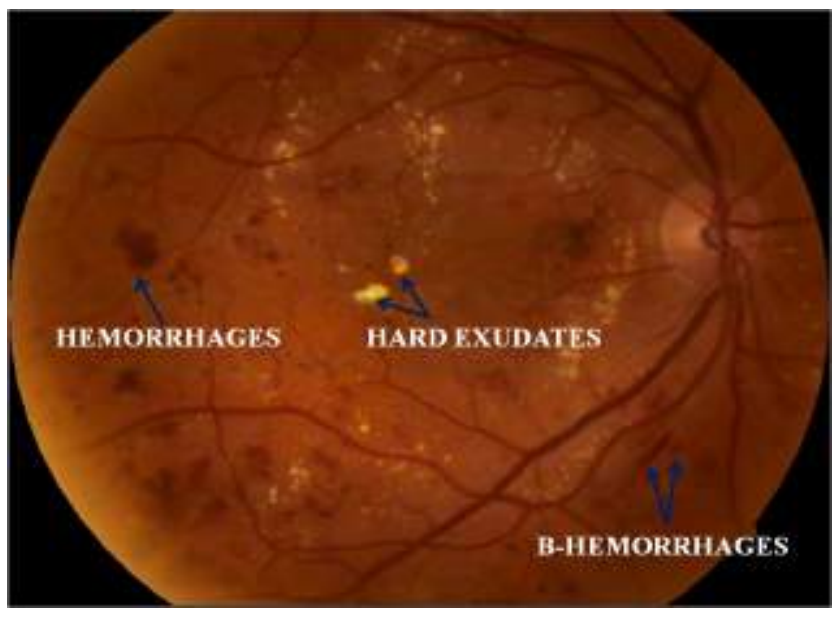

(b)

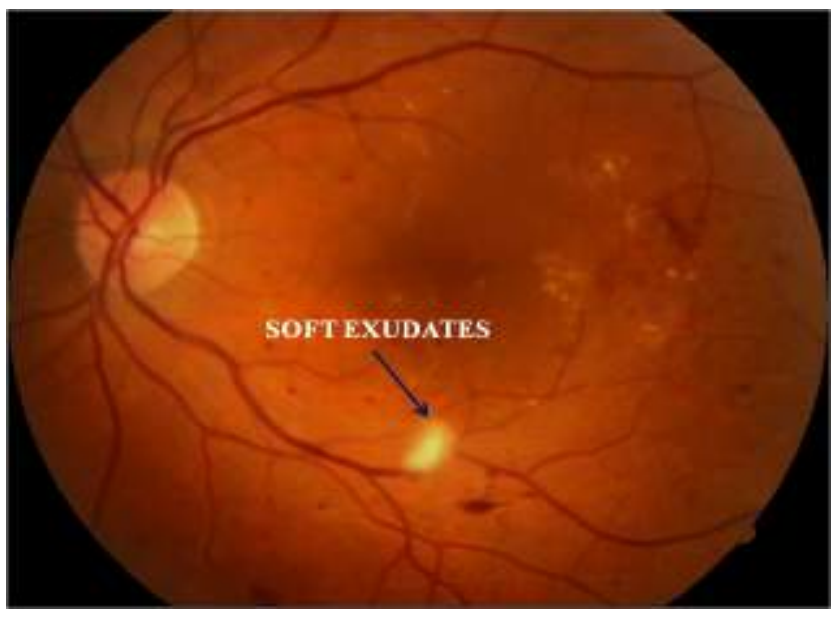

(c)

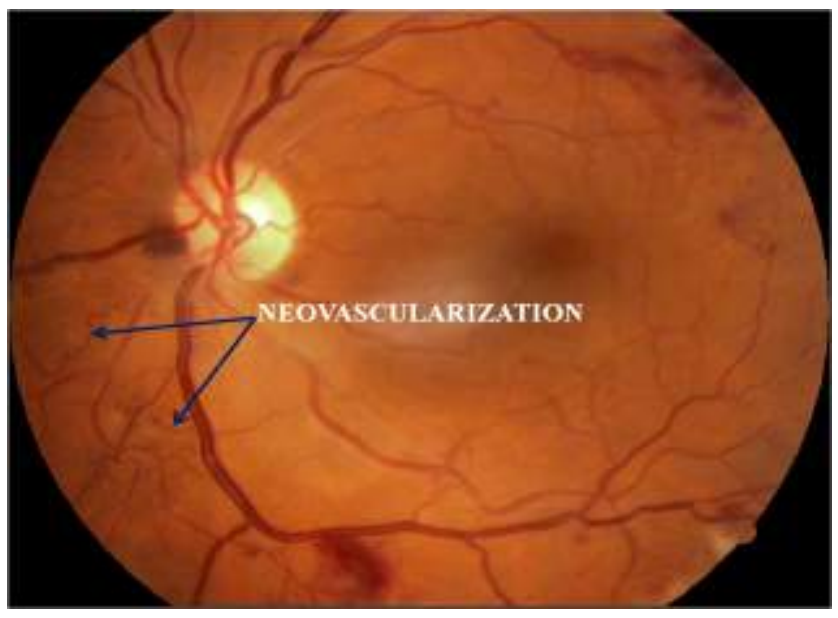

(d)

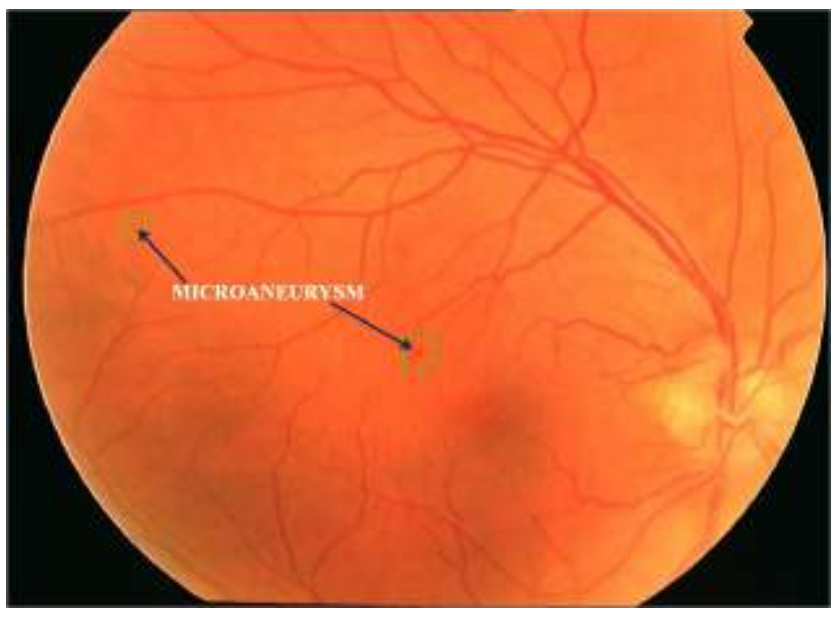

(e)

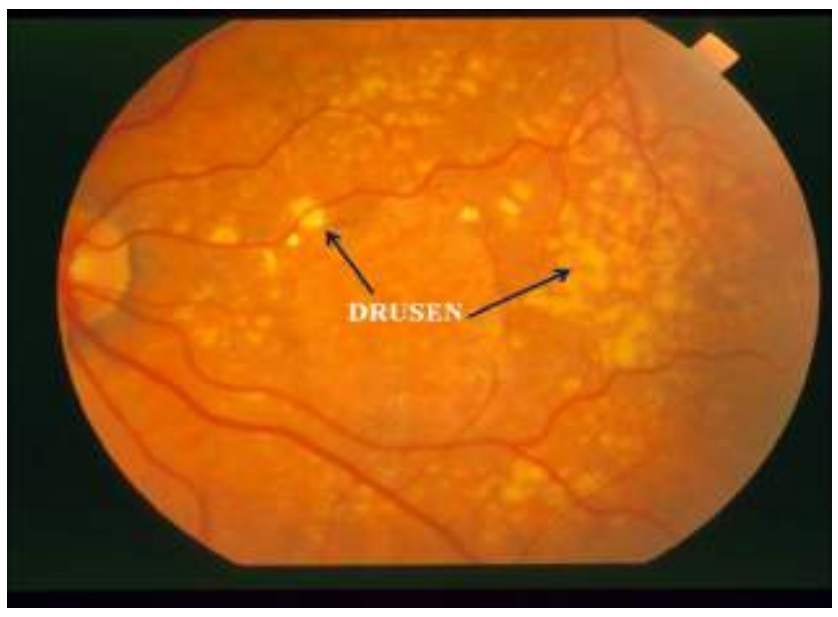

(f)

Figure 2.6: Typical fundus images; (a) normal, (b) hemorrhages, and hard Exudates (c) soft exudates, (d) neovascularization, (e) microaneurysms, (f) drusen.

Neovascularization, is the abnormal growth of new blood vessels on the inner surface of the retina as shown in Fig. 2.6(d) in an attempt to compensate for the lack of oxygen supply where such areas of retina send signals to stimulate the growth of new blood vessels in order to reestablish the supply of oxygen. These blood vessels are fragile and tend to bleed into vitreous cavity. As a result, vision can be obscured.

## 2.4 Diabetic Eye Diseases

Diabetes mellitus (DM) is a major medical problem throughout the world, that can be identified as a chronic condition connected with the impaired metabolism of glucose as a result of insulin deficiency or its resistance. Diabetes causes a wide variety of long term systemic complications that can affect heart, blood vessels, nerves, eyes and kidneys [20]. The number of people with diabetes is projected to increase from 2.8% to 4.4% in the time span of $2000-2030$ [19], moreover it is the common cause of legal blindness in individual between 20 and 65 years of age in the united states. Diabetic retinopathy and diabetic macular edema are the most widespread ophthalmic complications caused by diabetes.

## 2.5 Diabetic Retinopathy

Diabetic retinopathy (DR) is a microvascular complication that causes damages to blood vessels supplying the retina and is a leading cause of severe vision loss in people with diabetes if not properly treated [21]. Usually, it damages retinas in both eyes. However, early diagnosis and proper treatment can minimize vision loss. Fig. 2.7(a) and (b) show two different scenes viewed by a normal person and a person suffering from diabetic retinopathy respectively. DR can be classified into four stages as follows [23] :

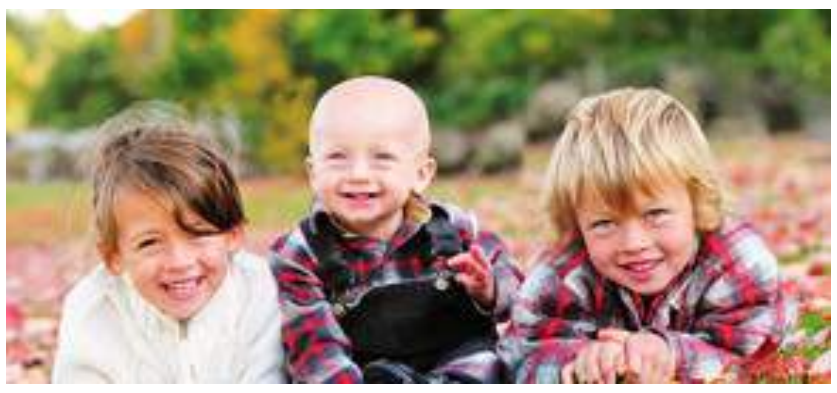

(a)

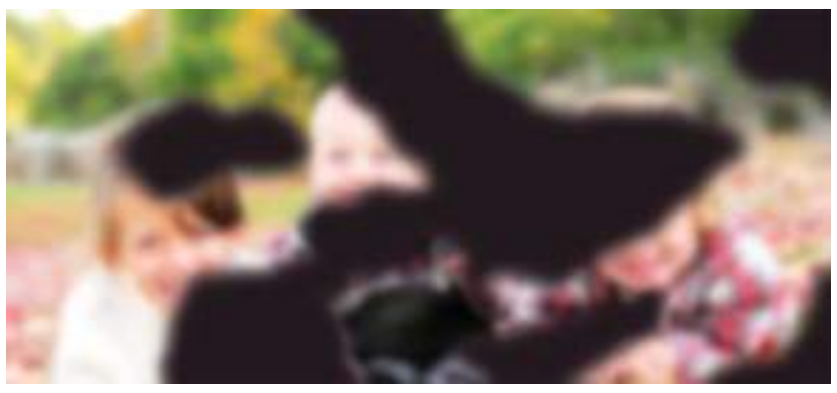

(b)

Figure 2.7: Normal vision and the same scene viewed by a person with diabetic retinopathy; (a) normal vision, (b) scene viewed by a person with diabetic retinopathy [22].

### 2.5.1 Mild Nonproliferative Retinopathy

The features of mild nonproliferative retinopathy are some of the earliest signs of diabetic retinopathy. At this point, microaneurysms occur along with hemorrhages and cotton wool spots. It is worth pointing out that not all patients will notice a change in their vision.

### 2.5.2 Moderate Nonproliferative Retinopathy

Many more microaneurysms, hemorrhages, and cotton wool spots appear, In addition to further damage to retinal blood vessels. Subsequently, the blood flow to the surrounding retinal tissue is reduced giving rise to vision loss.

### 2.5.3 Severe Nonproliferative Retinopathy

At this stage, large areas of the retina are deprived of blood flow. As a result, these areas of the retina send signal to the body in order to produce new blood vessels to enhance nourishment.

### 2.5.4 Proliferative Retinopathy

Proliferative retinopathy is the more advanced modes of diabetic retinopathy. At this advanced stage, new blood vessels start to grow as the retina sends signal to the body for nourishment enhancement. These new blood vessels are abnormal and fragile which grow along the retina and other parts of the eye. These blood vessels do not attribute symptoms or vision loss by themselves. However, their walls are thin and fragile, whenever they leak blood severe vision loss and even permanent blindness can occur.

### 2.5.5 Diabetic Macular Edema

Macular edema (ME) corresponds to accumulation of fluid and protein within the retina, specifically in the the outer plexiform and inner nuclear layers, as a non-distinct response to a breakdown in the blood retinal barriers. ME is the main reason of central vision loss in patients with diabetes mellitus and patients following intraocular operations [24]. According to [25] ME occurs if one of the following conditions is fulfilled:

1. There exists swelling/thickening of the retina including the center of the retina (macula) or the area within 500 µ of it.
2. Hard exudates present at or within 500 µ of the center of the retina with swelling of the adjacent retina.
3. If there is zone or zones of retinal swelling/thickening 1 disk area or larger in size; any part of which is within 1 disk diameter of the center of the macula.

### 2.5.6 Age related Macular Degeneration

Age related macular degeneration (AMD or ARMD) is a common eye condition that generally affects older adult and causes a loss of vision in the macula owing to damage of the retina. A blurred area near the macula is a frequent manifestation. In the long run, this blurred area may become larger, or blank spots may be developed in central vision. AMD occurs in three different forms depending on the number and size of drusen under the retina.

1. **Early AMD**. Generally, people with early AMD don't have vision loss, and medium sized drusen are common signs.
2. **Intermediate AMD**. Intermediate AMD may cause some vision loss. Large drusen, pigment changes in the retina or both are typically found in people affected by intermediate AMD.
3. **Late AMD**. There are two types of late AMD, Dry and wet. The former (geographic atrophy), is a progressive damage of the light sensitive cells in the macula which may lead to central vision loss. The latter (neovascular), is characterized by abnormal blood vessels growth underneath the retina that can leak fluid and blood, blurring or distorting central vision.

As a conclusion, this chapter summarized the human eye anatomy as well as the overall arrangement of retinal layers. It also discussed retinal imaging modalities, clinical lesions of the retina, and diabetic diseases. The next chapter reviews the literature regarding different methods used to discriminate between drusen and exudates.

# Chapter 3

# State of The Art

The growing prevalence of diabetes worldwide increases the number of cases that need to be reviewed by ophthalmologists. Additionally, the high cost of physical examination and lack of professional experts prevent a lot of people from receiving an adequate treatment. Computer aided detection (CAD) systems of retinal lesions associated with diabetes can offer many interesting benefits both in screening and clinical setting. In the former, it can give the opportunity to investigate a large number of images. In the latter, typical examination costs can be reduced as a consequence of reducing the workload of trained graders. In literature, a wide variety of CAD systems to detect retinal features and lesions involve three main steps. The first step is the preprocessing in order to compensate for great variability between and within retinal images. Green channel is considered the most preferable choice, because it provides a maximum contrast between different retinal lesions and structures. The second step is to extract candidate lesions, in some approaches feature selection procedure may be performed to remove redundant features. The last step is to classify candidate lesions into normal or abnormal. Table 3.1 summarizes different approaches used to classify images with bright lesions such as drusen and exudates in terms of preprocessing, methodology, features, classification, the size of the dataset, performance measurements (sensitivity, specificity, accuracy, and area under the curve).

This chapter is organized as follows: Section 3.1 discusses some mechanisms to detect and classify exudates, Section 3.2 discusses a few schemes in order to detect and classify drusen, while the last Section 3.3 will deliver some discussion about the different methods used to differentiate between drusen and exudates.

## 3.1 Automatic Detection and Classification of Exudates

The paper introduced by García et al. [26] is based on a neural network to extract hard exudates from color fundus images. The proposed method composed of prepossessing, segmentation, and classification uses three neural network classifiers including multilayer back-propagation (MLP), radial basis function (RBF), and support vector machine (SVM). Luminosity and contrast normalization have been performed on the green channel as a preprocessing step. Next, the enhanced image is subjected to a combination of global and local adaptive histogram thresholding in order to find candidate exudate regions that have to be classified later. Prior to classification the optic disk is removed since its characteristics are similar to exudate lesions. A total of 117 images are used, the training set consists of 50 images. After segmentation a number of regions are extracted, then given labels by experts as exudate or non exudate. In this way, they have created a fully labeled ground truth dataset to train the classifiers with these 50 images. The test set contains 67 images (27 are healthy retinas and 40 are DR patients). They have selected 18 features including (1-6) mean and std of the RGB values inside the region, (7-12) mean and std of the RGB values of the pixels belonging to a rectangular area around the region (5 pixels distance), (13-15) RGB values of the region, (16) region size; the number of pixels inside the region, (17) region compactness; the ratio between the square of the perimeter and the area of the region, in this situation a boundary tracking algorithm is needed, (18) region edge strength; the average of edge values in the perimeter of the region, in such case Prewitt operator is required. Ten fold cross validation is conducted to assess the generalization ability of the network, as well as two criteria are followed for final classification; image based criterion which is the ability of the algorithm to separate pathological images from normal images. lesion based criterion which is the number of exudates in the images that are correctly detected. Using a lesion-based criterion, they obtained a mean sensitivity (SE) of 88.14% and a mean positive predictive value (PPV) of 80.72% for MLP. With RBF they achieved SE=88.49% and PPV=77.41%, while they reached SE=87.61% and PPV=83.51% using SVM. With an image based criterion, a mean sensitivity (SE) of 100%, a mean specificity (SP) of 92.59% and a mean accuracy (AC) of 97.01% were achieved with MLP. Using RBF they reached SE=100%, SP=81.48% and AC=92.54%. With the SVM the image based results were SE=100%, SP=77.78% and AC=91.04%.

A detection and classification approach of diabetic macular edema (DME) severity is introduced by Deepak et al. [27]. HE can be considered as a standard manifestation to assess DME, where the severity of the risk is evaluated on the basis of the proximity of the HE to the macula. The detection part is achieved through a supervised learning approach on normal images, in such case any deviation from the normal characteristics might be an indication of abnormality. The center of the macula is automatically detected, after that the region of interest (ROI) is extracted by fitting the best circle within the fundus mask with macula at the center, moreover the optic is removed. They have proposed to generate motion patterns within ROI since HE may cause a bright smear pattern, whereas the textured background will be smoothed out. In order to create a motion pattern image $I_{MP}$, rotated versions of a given input image are generated, then these rotated images are combined using a mean or a maximum function to fuse all intensities at each pixel location. Radon based features are extracted from the image $I_{MP}$ at different angles and the desired feature vector then is constructed by concatenating the responses for different orientations. Subsequently, using a single class classifier each image can be classified as normal or abnormal hard exudate image. Disease severity is assessed via a rotational asymmetry metric, such that in normal cases the macula is relatively darker than other regions and characterized by a rotational symmetry. However, abnormal cases show asymmetry characteristics around the macula. Within the ROI eight angular samples were used to create eight patches and a histogram of 10 bins is created for each patch which is called the symmetry measure. Finally, at a given threshold, each ROI can be assessed as moderate or severe risk of DME. They have used images from four different publicly available datasets as follows: HEI-MED[1], MESSIDOR[2], Diaretdb[3], and Diaretdb1[4]. In total, they are 644 images of which 367 are normal and 277 are abnormal images. Ten fold cross validation is performed in order to ensure the result of DME assessment. The detection sensitivity is 100% with specificity between 74% and 90%. Immediate referral cases are detected with a sensitivity of 100% and specificity of 97%. The severity classification accuracy is 81% for the moderate cases and 100% for severe cases.

An automatic method for identification of retinal HE in color fundus images is proposed by Osareh et al. [28]. The proposed preprocessing combines two steps; image normalization via histogram specification and local contrast enhancement. The idea is to select a reference image, then adjust the values of each image such that its frequency histogram matches the reference image distribution. In the next step, local contrast enhancement is applied in order to distribute the values of pixels around the local mean. This operation is performed on the intensity channel after converting RGB images into HSI color space. The preprocessed images are segmented by fuzzy C-means (FCM) clustering algorithm. A total of 18 features are extracted in the same manner as [26], but with different color spaces. In the final stage, a three layer perceptron neural network is used for the classification purpose with 18 input nodes (selected feature space), 15 hidden units, and a single output node. They have used 75 color images, including 25 normal and 50 abnormal images for training. The output of the segmentation contains a number of exudate and non exudate regions, which are labeled by medical experts to obtain a fully labeled training dataset. In order to investigate the diagnostic performance of the system, 67 color images were used. The proposed system is able to achieve 95.0% sensitivity and 88.9% specificity for image based classification, and 93.0% sensitivity and 94.1% specificity regarding lesion based classification.

The method proposed by Giancardo et al. [29] for the DME diagnosis is based on the classification of a single feature vector per image, where this feature vector is found on three types of analysis: the exudate probability map, the color, and wavelet analysis. In the first type of analysis, the green channel besides the intensity channel from the HSI color space are employed. The images are resized to a predefined height with maintaining height/width ratio, then the background image is estimated by a large median filter whose size is approximately 1/30 the height of the fundus image. The normalization is enhanced with morphological reconstruction in order to improve the removal of nerve fiber layer and other structures at the edges of the optic nerve. The exudates probability map is obtained by a hard thresholding. The field of view (the black area around the fundus image) is identified through a fast method based on a region growing with four seeds placed at the corners of a downsampled version of the image, moreover the optic disk is removed as common to all methods in the literature. The exudate candidates are selected by running a kirsch compass kernel on the probability map and assigning a score for each candidate. In the second type of analysis, the colors of a new input image are equalized to a reference image by taking into account the scalar mean and standard deviation of that image. In the final stage, a stationary Haar wavelet analysis is performed up to the second level of the intensity channel of HSI color space. Color and shape features are extracted such as Avg, Std, Max, Min and Med from different color spaces via two different approaches; in the former, the exudate probability map is converted to two binary masks which are superimposed on the color and wavelet analysis outputs; in the latter, the exudate probability map is used to weigh the analysis outputs at a pixel level. Four different classifiers are used to evaluate the system performance. However, SVM with linear kernel achieves the best performance. The author provided a new publicly available dataset (HIE-MED) which consists of 169 patients from various ethnic groups and levels of DME, applied this one and another two publicly available datasets (MESSIDOR and DIARETDB1) they were able to achieve an area under the curve (AUC) between 0.88 and 0.94 depending on the dataset/features used.

[1]The Hamilton eye institute macular edema dataset (`http://vibot.u-bourgogne.fr/luca/heimed.php`)
[2]Methods to evaluate segmentation and indexing techniques in the field of retinal ophthalmology dataset (`http://messidor.crihan.fr/index-en.php`)
[3]Standard diabetic retinopathy database - calibration level 0 (`http://www2.it.lut.fi/project/imageret/diaretdb0/`)
[4]Diabetic retinopathy database and evaluation protocol (`http://www2.it.lut.fi/project/imageret/diaretdb1/`)

## 3.2 Automatic Detection and Classification of Drusen

Hijazi et al. [30] come up with an idea to classify retinal images as either age related macular degeneration (ARMD) or non ARMD by adopting a case based reasoning approach (CBR). CBR consists in three stages as follows:

**1- Image preprocessing**

This stage includes two steps; image enhancement and segmentation of retinal structures. In the former, color normalization is applied first, followed by illumination normalization and then contrast enhancement to increase the visibility of the main retinal anatomy. In the latter, 2D Gabor wavelet filters are applied to identify whether a pixel is classified as a vessel or non vessel by means of a Bayesian classifier.

**2- Spatial histogram generation**

Usually, color histogram is used as a simple way to represent an image for the purpose of object identification. In this approach, the number of colors per image is quantized (reduced) in order to reduce the computational time cost, and the image is partitioned into a number of equally sized regions. Subsequently, spatial histograms which keep the color and spatial information of the image are computed for each region. Then, all these histograms are concatenated into a single feature histogram. To sum up, each retinal image is represented as a series of histograms each encapsulated as a time series curve (the x-axis represents the histogram bins and the y-axis represents the number of pixels contained in each bin).

**3- Feature selection**

Basically, feature selection is a process to reduce the number of features by removing irrelevant (redundant) features, a class separability based method is adopted (Kullback-Leibler distance measure) for this reason. Finally, new cases are classified according to the most similar case in the case base (CB), since the histogram in CB can be represented as a time series curve a similarity measure can be achieved via a dynamic time wrapping (DTW). In order to measure the system performance, a tenfold cross validation method is applied; in addition, two parameters have been introduced including number of bins and the T parameter (Number of selected features). Their best results are conducted with a few number of color bins (32) and a T parameter of 5 where specificity, sensitivity, and accuracy are 74%, 79%, and 77% respectively for a total of 144 retinal images described as 86 AMD and 58 non AMD images.

Hijazi et al. [31] proposed another approach which relies on image angular and circular decomposition. The output of the decomposition step is a set of tree represented images (each image is represented as a tree). A weighted frequent sub-tree mining approach is used to determine the most often occurring sub-trees. The weighted frequent sub-trees are then employed to adapt the training input data in a vector representation form (one vector per image) through a linear support vector machine to reduce the dimensionality of the selected features. Two classifiers are then used for final classification, support vector machine and Naive Bayes. A total of 258 images from two different publicly dataset, i.e. ARIA[5] and STARE[6] including 160 AMD and 98 normal are used. The classifier's performance is assessed through a class classification using tenfold cross validation. They have achieved an accuracy of 100% with SVM and slightly similar result (95% accuracy) with Naive Bayes.

Akram et al. [32] proposed an algorithm to automatically segment drusen in fundus images for ARMD diagnosis. The proposed algorithm incorporates three steps. Firstly, dark regions are eliminated using morphological closing, after that an image intensity enhancement is performed by applying an adaptive histogram equalization. Secondly, they have adopted Gabor kernel based filter banks in order to find all possible bright lesions (candidate selection). Optic disk is eliminated using Hough transform. Finally, some features have been computed, i.e. area, compactness, average boundary intensity, minimum boundary intensity, maximum boundary intensity, mean hue, mean saturation, mean intensity value, mean gradient magnitude. Least square support vector machine is adopted to perform the classification. They have used images from the STARE dataset, the dataset encompasses 400 images out of these 58 images contains drusen. The accuracy, sensitivity, and specificity achieved are respectively 97%, 95% and 98.4% . These parameters are computed by comparing the proposed system with ground truth data. However, the system achieved 100% accuracy in finding drusen at image level.

Zheng et al. [33] introduced a system which combines a set of algorithms i.e. pattern recognition, computer vision, and machine learning. Image preprocessing incorporates several processes. These encompass image denoising (non local mean filtering), retina mask generation (image thresholding and some morphological erosion), illumination correction, and color transfer to return all test images similar in color. The detection strategy is found on two consecutive procedures, i.e. a pixel wise classification and a group wise classification. The idea of the former, is to detect whether a pixel is a drusen or not using color image descriptors (Hessian features) and multi-scale image local descriptors (total variation features). They have used Ada-boost for feature selection and Least square support vector machine for classification. The group wise classification is exploited to remove false positive components from pixel wise classification and it is accomplished directly by Least square support vector machine. The system's validation is made by comparing its output to manually segmented drusen on a pixel by pixel basis. Two different datasets are used as follows: 50 images from CAPT[7] and 88 from AMISH. Accuracy between 80% and 86%, sensitivity between 82% and 87%, and specificity between 71% and 78% are respectively achieved based on the dataset.

[5] Automated retinal image analysis dataset (`http://www.eyecharity.com/aria_online.html`)
[6] Structured analysis of the retina dataset (`http://www.ces.clemson.edu/~ahoover/stare/`)
[7] The complications of age-related macular degeneration prevention trial dataset

## 3.3 Automatic Detection and Discrimination of Drusen and Exudates

Niemeijer et al. [6] developed an algorithm that can automatically detect bright lesions in retinal images and can differentiate among exudates, cotton wool spots, and drusen. One hundred thirty images containing all bright lesions are used to build the training set. All pixels in these images are segmented by retinal specialists whether they are exudates, cotton-wool spot, drusen or background retina. Vessels, optic disc, and red lesions were treated as background retina. Three hundred images are used to perform diagnosis validation (100 images contain lesions and 200 images contain no lesions). The overall methodology is composed of four stages; probability map generation, bright lesion pixel clusters, bright lesion detection, and bright lesion classification. Firstly, the green channel is convolved with a set of 14 digital filters, because bright lesion filter responses will be different from the filter responses for non lesion pixels. Secondly, a K-nearest neighbor (KNN) classifier is used to classify the pixels based on the filter responses such that pixels with a probability higher than a predetermined threshold are grouped into bright lesion pixel clusters. Each bright lesion pixel cluster is usually referred to as a potential lesion. Thirdly, a second KNN classifier is trained using a set of sample potential lesions extracted from the training set to discard false bright lesion clusters. Each bright lesion pixel cluster is assigned a probability indicating the likelihood that the pixel cluster is a true bright lesion based on a number of features like the contrast of clusters in the RGB color space, in addition to size, shape, and contrast of potential lesions; their proximity to the closest vessel; and proximity to the closest red lesions, where the more closer potential lesions from red lesions the more likely to be true bright lesions. Finally, a linear discriminant analysis classifier is trained using the number of red lesions in the image, the number of detected bright lesions, and the bright lesion cluster probability to classify the bright lesions into exudates, cotton wool spots, or drusen. The system achieved an area under the curve, sensitivity, and specificity of 95%, 95%, and 88% for any type of bright lesion detection, $95\% - 86\%$, $70\% - 93\%$, and $77\% - 88\%$ for exudates, cotton-wool spots, and drusen, detection.

Grinsven et al. [34] provided an algorithm to automatically discriminate and retrieve images with exudates or drusen. The algorithm initializes with partitioning the image (after mean subtraction of the color planes) into a fixed number of squared patches. The set of features appropriate for discriminating between normal and abnormal are as follows: color histogram features obtained from the RGB, HSV and YCbCr color planes: R and B planes (32 bins each); S (64 bins) and V (32 bins); Cb (16 bins). Histogram of Laplacian of Gaussian (8 bins, $sigma = 12$ of Gaussian), HoG (64 orientation bins); LBP(16 bins) and granulometry (scales 1,3,5,7,9,11). While, the set of features suitable for differentiating between drusen and exudates are such as: color histogram features extracted from the green and blue color planes (16 bins each); H, S planes (64 bins each) and V (32); Y (32). HoG and LBP (128 bins each). By using Ada-Boost features selection technique, the two sets of features are reduced to 24 and 32 dimensional feature vectors respectively. Features are selected empirically by adopting a class validation with 5 fold cross validation. A bag of words approach is then utilized for two purposes image retrieval as well as image classification. Some similarity measures are employed to help retrieve images, i.e., squared chord, L1-norm, L2-norm. However, the best result is achieved using weighted squared chord, while a weighted nearest neighbor approach is used for the classification using the distance as a similarity metric. Three different datasets including STARE, MESSIDOR, and EUGENDA[8] with a total of 415 images are employed. They have constructed two datasets: Set A and Set B. Each set is further subdivided into a training set and test set. The test set is similar to set A and B. In this manner, they can test how well the method generalizes when the test set contains only images from a different population. A precision of 0.76 and AUC of 0.9 are achieved respectively for retrieval and classification.

Deepak et al. [35] suggested to use a visual saliency based framework for bright lesion detection including hard exudates and drusen. The green channel of a given image is top hat filtered using a disk shaped structural element in order to enhance the local contrast. Then, the background image is estimated using median filtering and subtracted from the enhanced image to minimize the effect of uneven illumination. A contrast stretching is performed as a final step in preprocessing prior to saliency computation. The spectral residual (SR) model is employed for saliency computation in order to detect perceptual objects i.e. hard exudates and drusen. They have used the generalized motion patterns (GMP) they have proposed in [27]. Feature extraction is performed on the saliency map using GMP by extracting radon features with orientations between $0^\circ - 180^\circ$ with a step of $3^\circ$ and 256 bins. At this stage images are classified as normal or having lesions using a k-nearest neighbor (K-NN) classifier. The saliency map provides useful information on locations of abnormalities in an image, so it may help differentiate between an image with hard exudates and drusen. The saliency image is divided into a fixed number of square patches. Patches with lesions are identified if their saliency value is greater than a predetermined threshold. Local binary patterns (LBP) are extracted from the green channel of all patches with lesions, then they are fed to a support vector machine (SVM) classifier with RBF kernel in such case every abnormal image can be classified as having HE or drusen. They have used five publicly available dataset (HEI-MED, Diaretdb, MESSIDOR, ARIA, STARE), an area under the curve of 0.88 to 0.98 for detection and accuracies ranging from 0.93 to 0.96 for lesion discrimination are yielded based on the dataset used.

So far, we have summarized the different techniques used in the literature to classify images with bright lesions, namely drusen and exudates as shown in Table 3.1. The proposed methodology will be elaborated in depth in the next chapter.

[8]The European genetic dataset (`http://www.eugenda.org/`)

| Authors | Lesions | Preprocessing | Methodology [State] | Features {L} | Classification | Dataset | | Results | | | |
|---|---|---|---|---|---|---|---|---|---|---|---|
| | | | | | | Train | Test | SE% | SP% | AC% | AUC |
| García et al. [26] | EX | Luminosity normalization & contrast enhancement | Global and local histogram thresholding [✗] | Color and shape features {18} | NNs (10 FCV) | 50 | 67 | 100 | 92.59 | 97.01 | ... |
| Deepak et al. [27] | EX | Not necessary | Create motion patterns of a circular ROI [✓] | Radon based features {36} | PCA (10 FCV) | 367 | 277 | ... | ... | ... | 0.92 |
| Osareh et al. [28] | EX | Histogram specification and contrast enhancement | FCM clustering [✗] | Color and shape features {18} | NNs | 75 | 25 | 93 | 94.1 | ... | ... |
| Giancardo et al. [29] | EX | Image normalization | Color wavelet decomposition [✓] | Statistical features (Avg, Std, Max, Min, Med) {48} | SVM (LOOCV) | 169 | ... | ... | ... | ... | 0.94 |
| Hijazi et al. [30] | Dru | Luminosity normalization & contrast enhancement | 2D Gabor wavelet filters [✗] | Spatial histograms {160} | DTW (10 FCV) | 144 | ... | 79 | 74 | 77 | ... |
| Hijazi et al. [31] | Dru | Luminosity normalization & contrast enhancement | Angular and circular decomposition [✓] | weighted frequent sub-graph mining approach {3671} | SVM (10 FCV) | 258 | ... | 100 | 100 | 100 | ... |
| Akram et al. [32]* | Dru | Morphological closing and histogram equalization | Gabor filter banks [✗] | Color and shape features {10} | LS-SVM | 400 | ... | 95 | 98.4 | 97 | ... |
| Zheng et al. [33]* | Dru | Mean filtering, retinal mask generation and illumination correction | Pixel and group based classification [✓] | Hessian and total variation features | LS-SVM | 88 | ... | 87 | 78 | 86 | ... |
| Niemeijer et al. [6] | Dru & EX | Convolution with 14 digital filters | bright lesions clustering KNN [✗] | Shape, color, contrast, and position {83} | LDA | 130 | 300 | 95 | 88 | ... | 0.95 |
| Grinsven et al. [34] | Dru & EX | Mean subtraction | Bag of Words approach [✓] | Histograms, LOG, HOG, LBP, and granulometry {58 × no of patches} | KNN (5 FCV) | 225/379 | 36 | ... | ... | ... | 0.9 |
| Deepak et al. [35] | Dru & EX | Top hat filtering, in addition to contrast stretching | Visual saliency map and GMP [✓] | LBP {256 × no of patches } | SVM (10 FCV) | 388 | ... | ... | ... | ... | 0.96 |

Table 3.1: Summary of several methods used to discriminate hard exudates, and drusen. This table shows results of the image based criterion. However, [.]* represents a lesion based criterion. State: the paper uses more than one dataset, L: the length of the feature, EX: Exudates, Dru: Drusen, SE: mean sensitivity, SP: mean specificity, AC: mean accuracy, AUC: Area under the curve, FCV: fold cross validation, LOOCV: leave one out cross validation, NNs:neural networks, KNN: K-nearest neighbour, SVM: support vector machine, LS: least square, LDA: linear discriminant analysis, DTW: dynamic time wrapping, GMP: generalized motion patterns.

# Chapter 4

# Methodology

The proposed method is based on the bag of words (BOW) approach to automatically discriminate between normal, drusen, and exudates in color retinal fundus images. In this approach, SURF as well as HOG and LBP features are extracted from local regions of retinal images. Then, a visual codebook is constructed using K-means clustering algorithm. The cluster centers are considered as visual words within the codebook. Each individual feature in the image is quantized to the nearest word in the codebook, and an entire image is substituted by a global histogram counting the number of occurrences of each word in the codebook. The size of the resultant histogram is the same as the number of words in the codebook and also the number of clusters obtained from the clustering algorithm. The Final histogram representation is fed into a linear kernel SVM for classification. Fig. 4.1 shows the flowchart of training phase, while Fig. 4.2 shows the flowchart of testing phase.

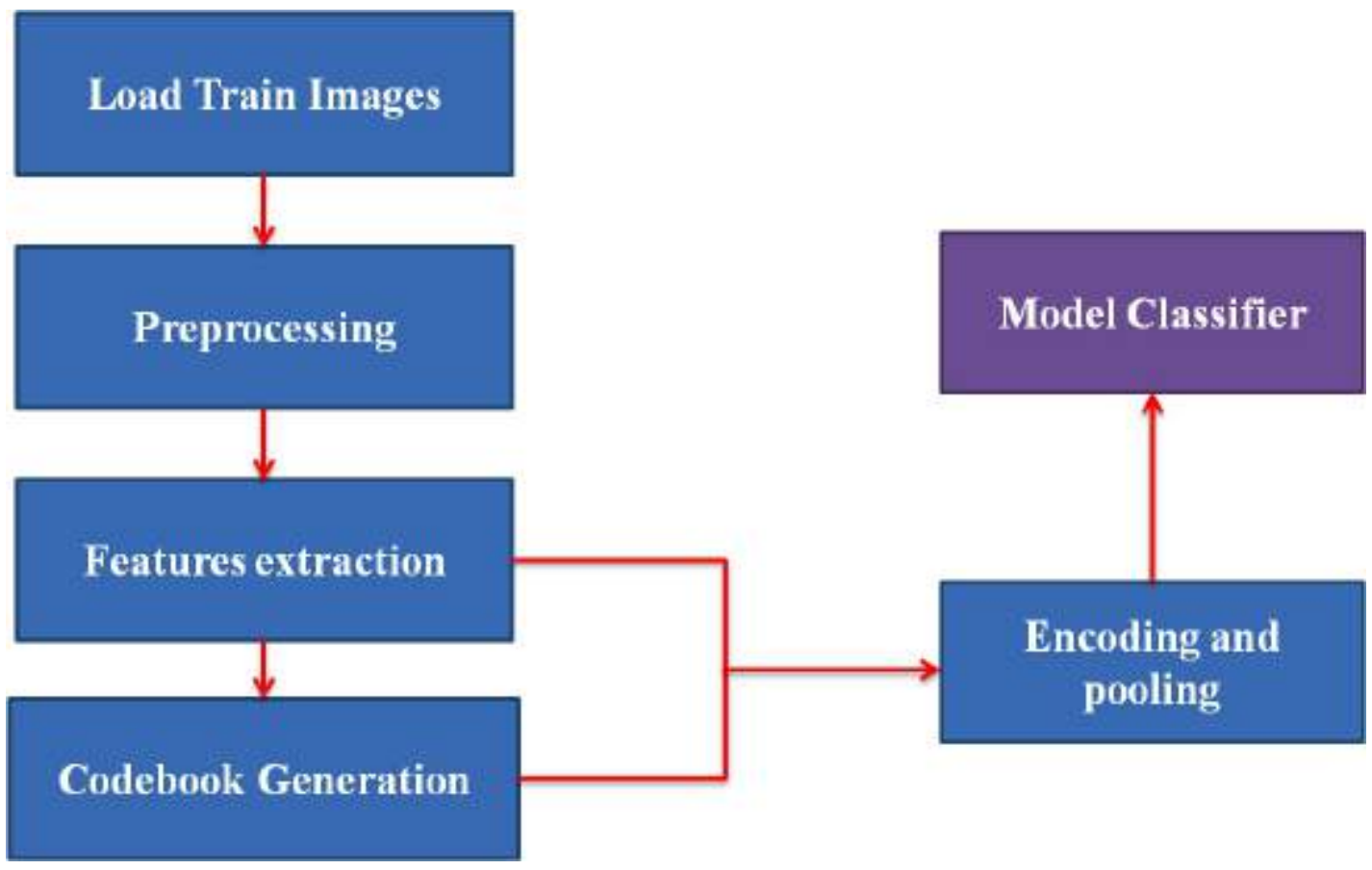

Figure 4.1: Training flowchart.

This chapter is organized as follows: Section 4.1 describes steps involved in the preprocessing step, Section 4.2 discusses in detail different features employed for abnormality detection, Section 4.3 describes the codebook construction using k-means clustering algorithm, Section 4.4 investigates the encoding method as well as the histogram pooling process, and Section 4.5 explains the linear support vector machine which is used for the final classification purpose.

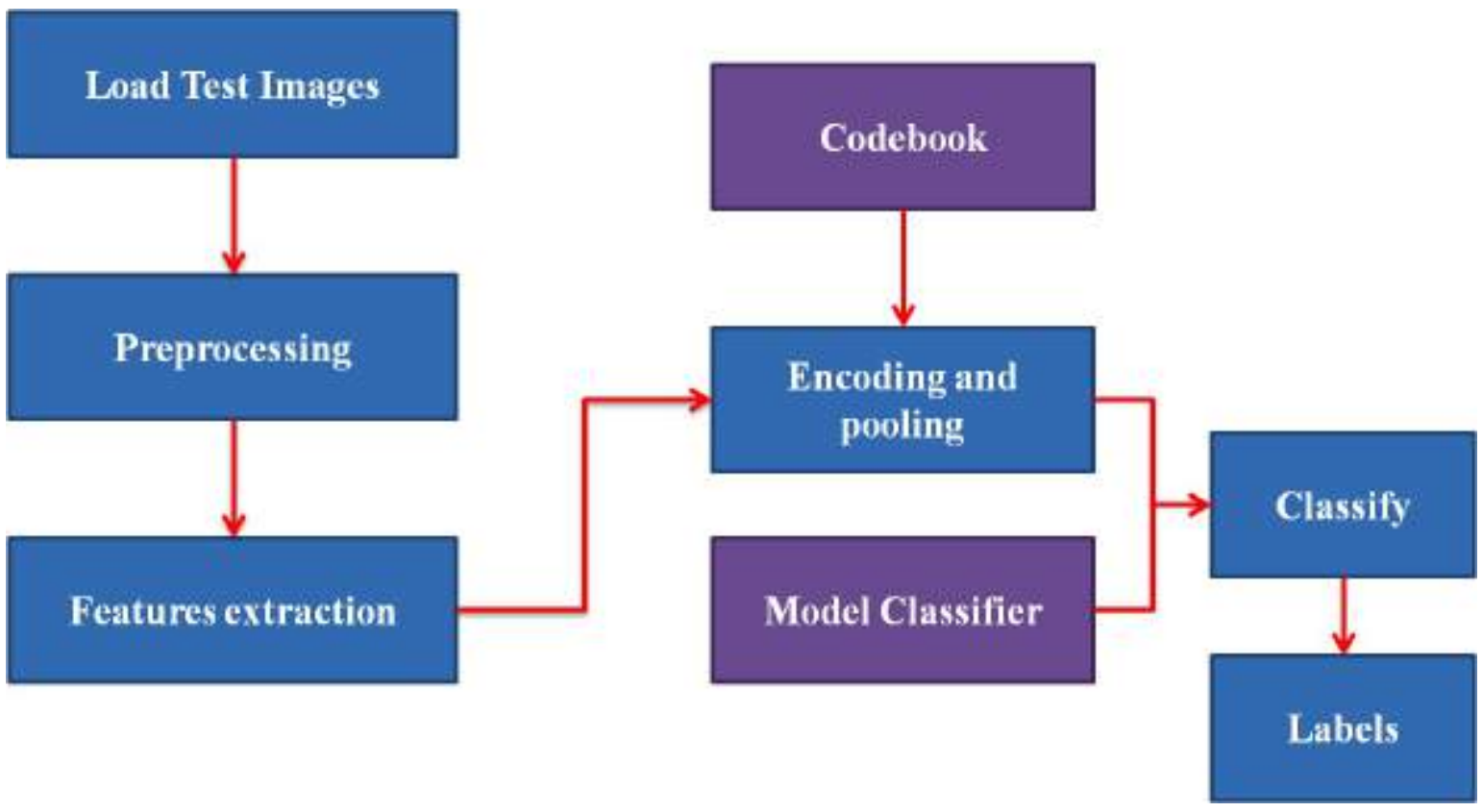

Figure 4.2: Testing flowchart.

## 4.1 Preprocessing

The goal of the preprocessing step is to compensate for image variation by normalizing the original image against a reference image. Typically, image variation appears within the same image (intra image variability) as well as between images (inter image variability). The reasons of intra image variability are differences in light diffusion, the presence of abnormalities, variation in reflectivity, and fundus thickness, while inter image variability is caused by differences in cameras, illumination, acquisition angle, and retinal pigmentation. In the RGB color space, the green channel provides the best contrast between the blood vessels and background whereas the red channel suffers from over−saturation and the blue channel can be under−saturated and noisy [36] as shown in Fig. 4.3.

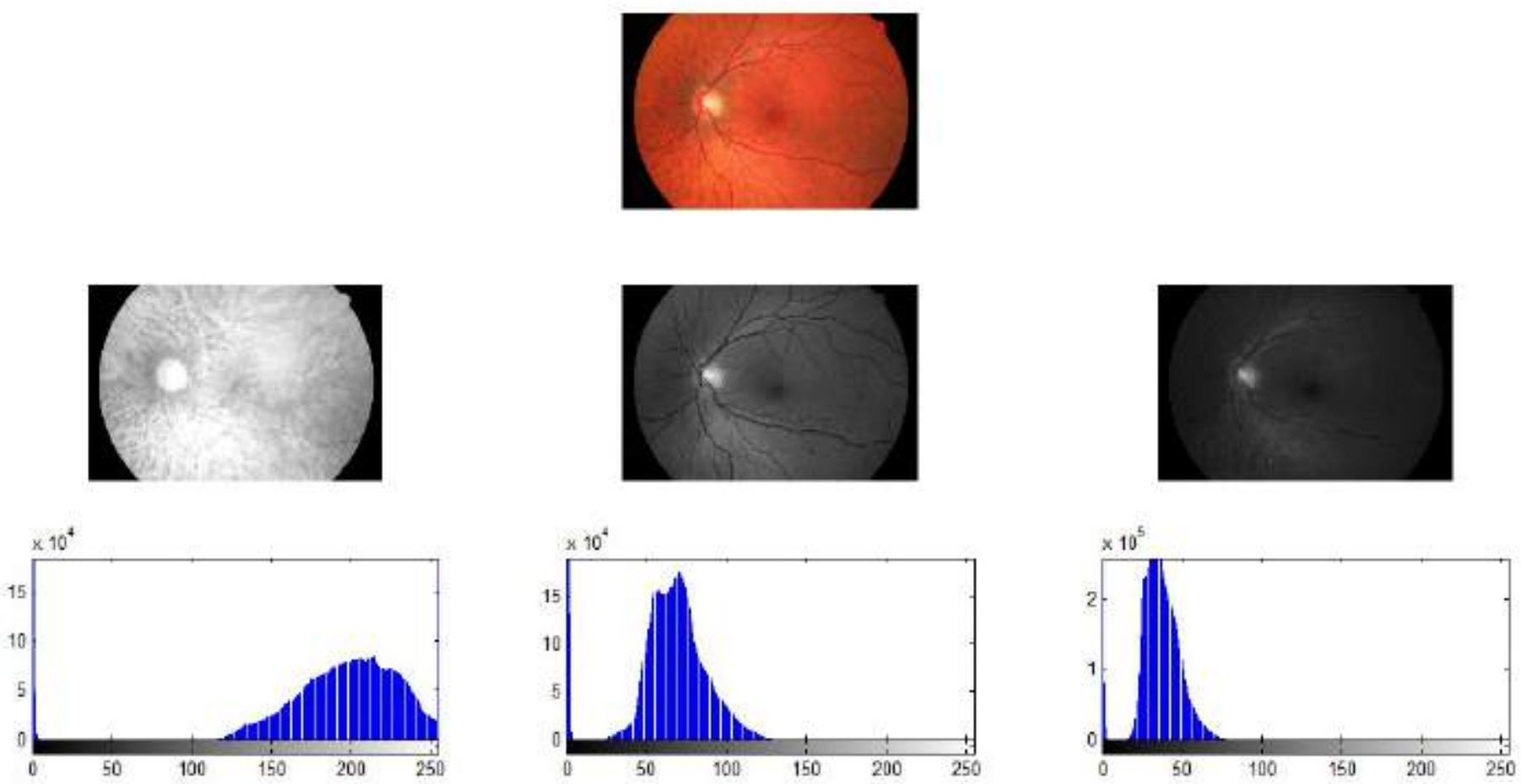

Figure 4.3: Intensity distribution of a sample color fundus image. The image's channels are respectively red, blue, and green from left to right.

### 4.1.1 Intensity normalization

The main purpose of the intensity normalization is to reduce the inter and intra patient variability. According to the paper introduced by Cree et al. [37] the background-less fundus image has normally distributed colors. Thus, the image can be represented by the scalar mean (μ) and standard deviation (σ) throughout the entire image. If these two parameters are calculated for a reference image, it is possible to equalize the colors of the new image to the reference one in a more effective manner than simple histogram equalization [29]. In this work, the mean (μ) and std (σ) are empirically chosen for all datasets instead of computing them from a reference image. Furthermore, the preprocessing is applied only to the green channel rather than the three planes of the RGB color space. All images are resized to a height of 512 pixels, while maintaining the aspect ratio because of their large sizes. The description of the process for a single color plane is explained as follows:

$$
\begin{aligned}
&\mu_{ref} = 0.5 \\
&\sigma_{ref} = 0.1 \\
&I_{out} = I_{in} - \text{medianFilter}(I_{in}) \\
&\mu_{out} = mean(I_{out}) \\
&\sigma_{out} = std(I_{out}) \\
&I^1_{out} = (I_{out} - \mu_{out}) \div \sigma_{out} \\
&I^2_{out} = (I^1_{out} \times \sigma_{ref}) + \mu_{ref}
\end{aligned}
\tag{4.1}
$$

The background image is estimated by a median filter, whose size is approximately $\frac{1}{30}$ the height of the fundus image. $I_{in}$ is the image to be equalized, $I_{out}$ is the background-less image, and $I^2_{out}$ is the equalized image. Fig. 4.4 shows an example for normal image equalization as well as drusen image equalization. Although the two images (Fig. 4.4 (a) and (b)) have different ethic backgrounds and quality level, the resultant (Fig. 4.4 (c) and (d)) images have very similar colors.

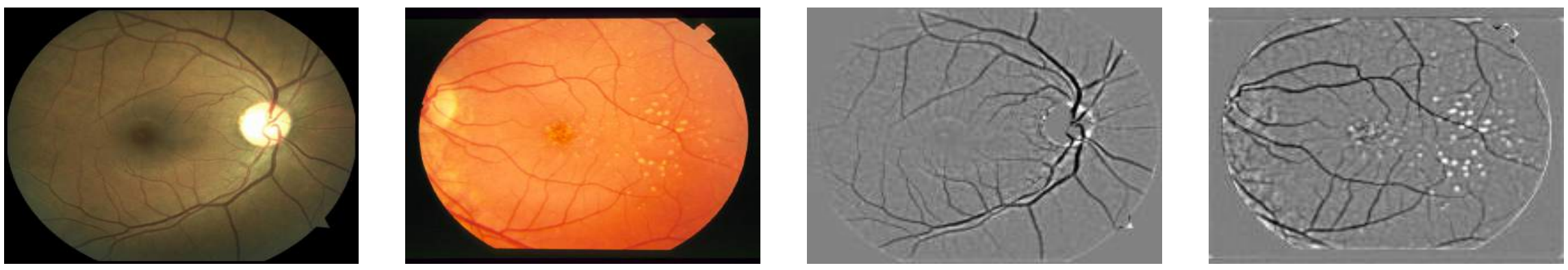

(a) (b) (c) (d)

Figure 4.4: (a) Normal image, (b) drusen image, (c) equalized normal image, and (d) equalized drusen image.

## 4.2 Feature Extraction

Feature extraction is nothing but transforming the input image into a set of features that efficiently represent interest parts of that image. Thus, these features should be carefully chosen. Based on different approaches revised in the literature [34, 35, 50], it is noticeable that local binary patterns (LBP), speeded up robust features (SURF), and histogram of oriented gradients (HOG) achieve satisfactory results when they are used for diabetic retinopathy diagnosis. In this work, all features are extracted from the three channels of the RGB color space. As discussed in the previous section the preprocessing is only applied to the green image, this is because in some datasets such as STARE if a preprocessing step is applied on the red or blue channel, most of the information would be probably lost as shown in Fig. 4.5.

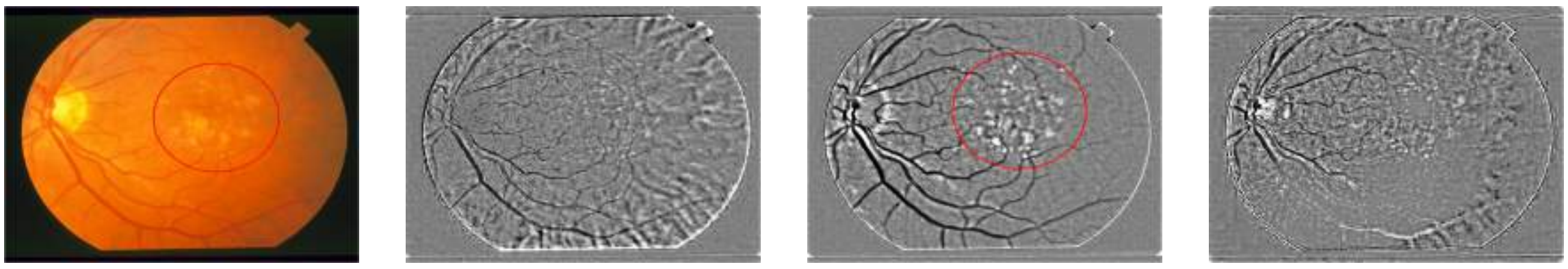

(a) (b) (c) (d)

Figure 4.5: Preprocessing result on a drusen image from the STARE dataset. (a) drusen image, (b) enhanced red image, (b) enhanced green image, and (d) enhanced blue image. In the red and blue image, the drusen (inside the red circle) cannot be identified as opposite to the green image.

Typically, The dimension of the SURF descriptors are 64 × number of interest points (see Section 4.2.1). In our case two approaches are adapted as follows: ordinary SURF and dense SURF. In the first, SURF descriptors are extracted from all RGB color channels, then they are horizontally concatenated to get a feature matrix of a size 64 × total number of interest points extracted from the three channels as shown in Fig. 4.6 (a) and (b). In the second, SURF descriptors are extracted from a dense grid uniformly distributed throughout the image i.e. SURF descriptors are computed on 16 × 16 pixel patches (non overlapping) with a spacing of 16 pixels, this is shown in Fig. 4.6 (c). As opposite to ordinary SURF, for each patch we get a feature vector of a dimension 64, then for each image channel we get a feature matrix of a size 64 × number of patches. Finally, each image constitutes a feature matrix of a size 192 × number of patches by vertically concatenating each feature matrix. The implementation of SURF is done using mat-lab built in function.

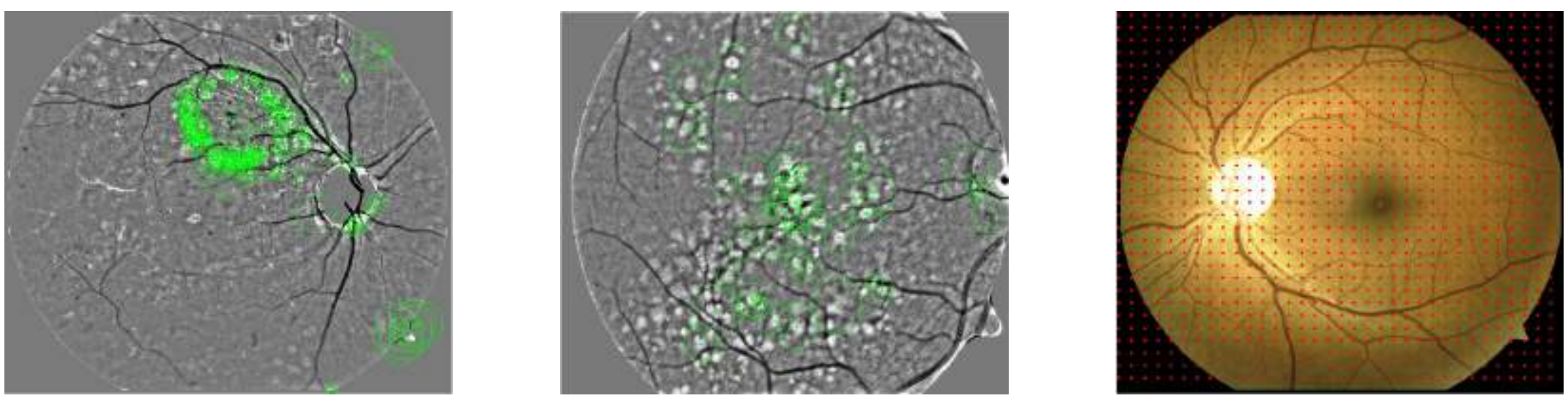

(a) (b) (c)

Figure 4.6: SURF and dense SURF interest points. (a) and (b) show SURF points on the green channel of an exudate and a drusen image respectively. (c) shows dense SURF points on a normal image.

The HOG descriptors are obtained as similar to [42] (see Section 4.2.2), due to its lower dimension and discriminative power. Each image channel is divided into fixed number of blocks with a size of 32 × 32 pixels, then each block is subdivided into 4 cells (each cell is 16 × 16 pixels), as a result each block contributes to a feature histogram of a dimension 31. For each channel, the histograms are vertically concatenated forming a feature matrix of a size 31 × number of blocks. In total, the three channels will constitute a feature matrix of a size 93 × number of blocks. The null descriptors originating from the black area surrounding the fundus image are not are not taken into account. Fig. 4.7 shows an example of HOG descriptors extraction.

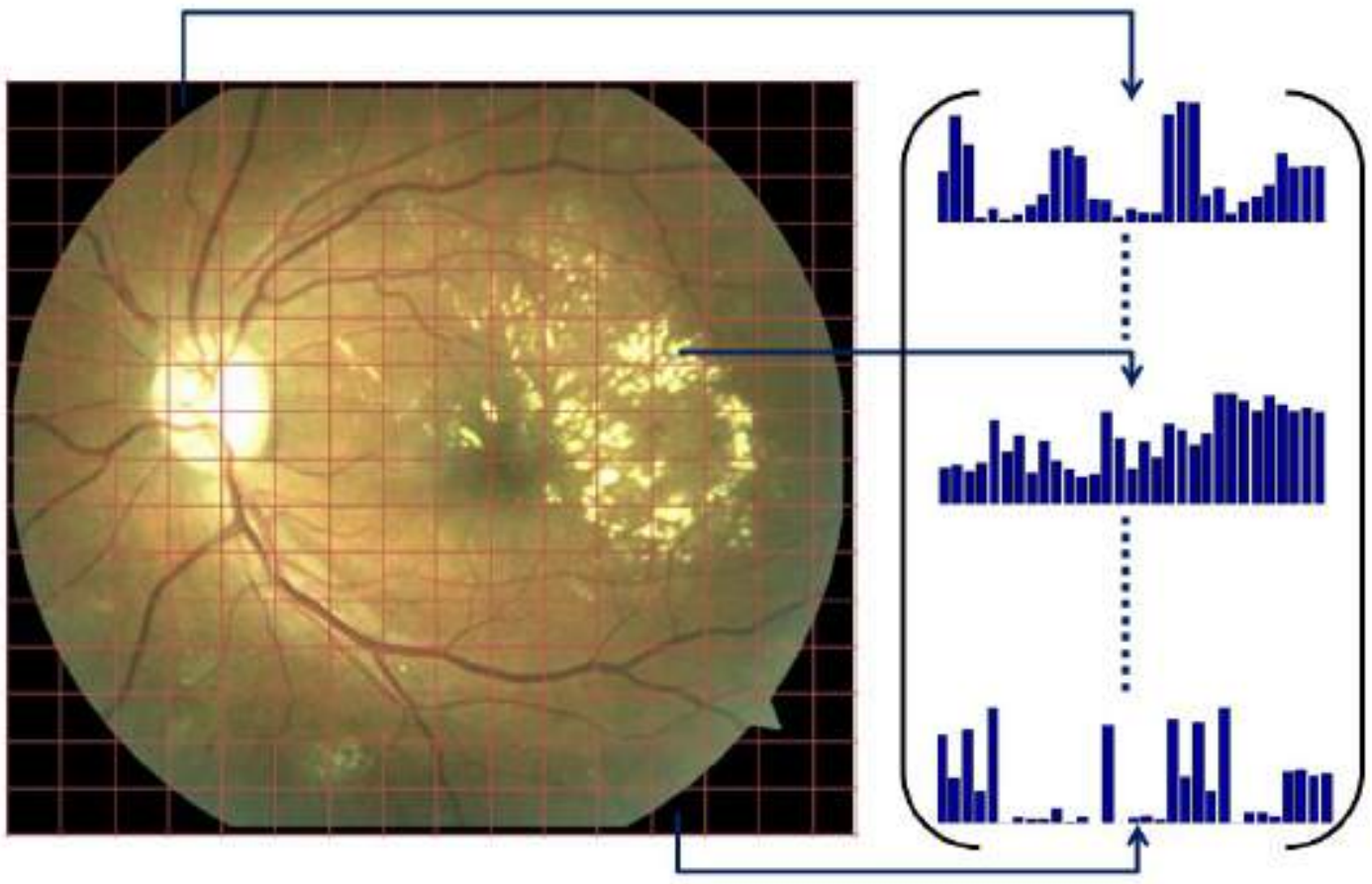

Figure 4.7: HOG descriptors. The image is divided into blocks and a histogram is execrated from each block.

The LBP descriptors are extracted from local patches of a size 32 × 32 pixels similar to the HOG (see Section 4.2.3). However, for each patch LBP features are computed using a 3 × 3 moving window centered at each pixel within the patch. The uniform local binary patterns are selected because of its lower dimension and reducing the number of codes inflicted by high frequency noise. The size of the feature matrix per image is 58 × number of patches, so in total we get a feature matrix of a size 174 × number of patches. The HOG and LBP are implemented using the VLFeat open source library [45].

### 4.2.1 Speeded Up Robust Features (SURF)

SURF (Speeded Up Robust Features) is a local invariant interest point detector−descriptor, first proposed by Bay et al. [38]. It is based on two steps: fast interest point detection and distinctive point description. In the previous, integral images are used since they allow for fast computation of box type convolution filters. Integral images can be created by Eq. (4.2).

$$ii(x, y) = \sum_{i=0}^{i \leq x} \sum_{j=0}^{i \leq y} i(x, y) \tag{4.2}$$

where $i(x, y)$ is the brightness values of the pixel in original image and $ii(x, y)$ is the integral image. Introducing integral image, the area of any upright rectangular region takes only three arithmetic operations i.e. calculation time is independent on its size. The sum of brightness in area $D$ is calculated as $S4 + S1 - (S2 + S3)$ as shown in Fig. 4.8.

The keypoint detection is based on the determinant of the hessian matrix. A point $x = [x, y]^T$ in original image $i$, the hessian matrix H(x,σ) in $x$ at a scale (σ) is defined as follows:

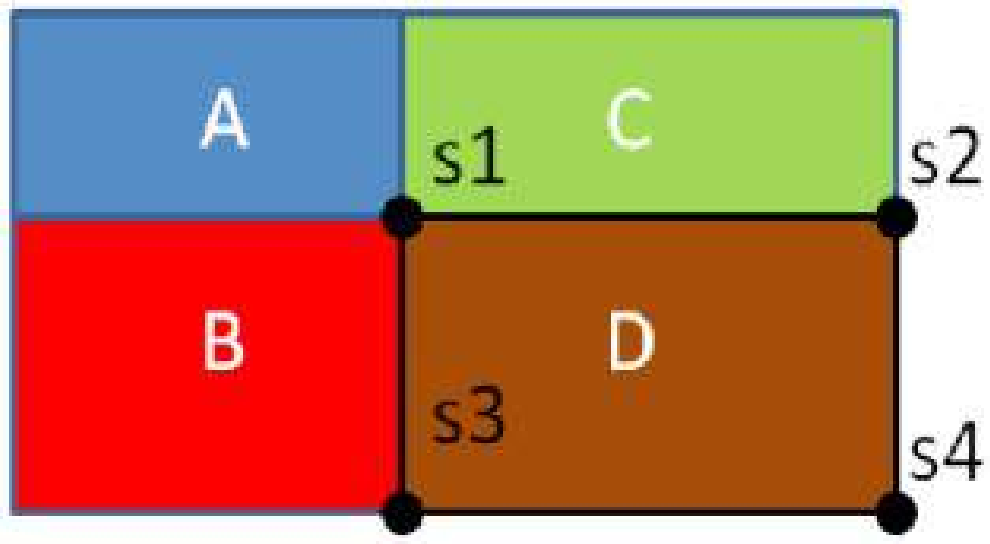

Figure 4.8: Integral image [39]. S1 is the sum of pixels in rectangle A, similarly S2, S3, and S4 show A+C, A+B, A+B+C+D respectively.

$$H(x, \sigma) = \begin{bmatrix} L_{xx}(x, \sigma) & L_{xy}(x, \sigma) \\ L_{xy}(x, \sigma) & L_{yy}(x, \sigma) \end{bmatrix} \tag{4.3}$$

Here $L_{xx}(x,\sigma)$ is the convolution of the second order Gaussian derivative with the image at a point $x = [x, y]^T$ and the same for $L_{yy}$ and $L_{xy}$. These derivatives are known as the Laplacian of Gaussian. SURF proposed an approximation to these derivatives by using box filter representation. $D_{xx}(x,\sigma)$ is an approximation to $L_{xx}(x,\sigma)$ likewise $D_{xy}(x,\sigma)$ is $L_{xy}$, Fig. 4.9 shows an example of box filter approximations. The performance increases when these filters are used in conjunction with the the integral image.

The approximation of the hessian determinant is given by Eq. (4.4).

$$det(H_{\text{approx}}) = D_{\text{xx}} D_{\text{yy}} - (0.9 D_{\text{xy}})^2 \tag{4.4}$$

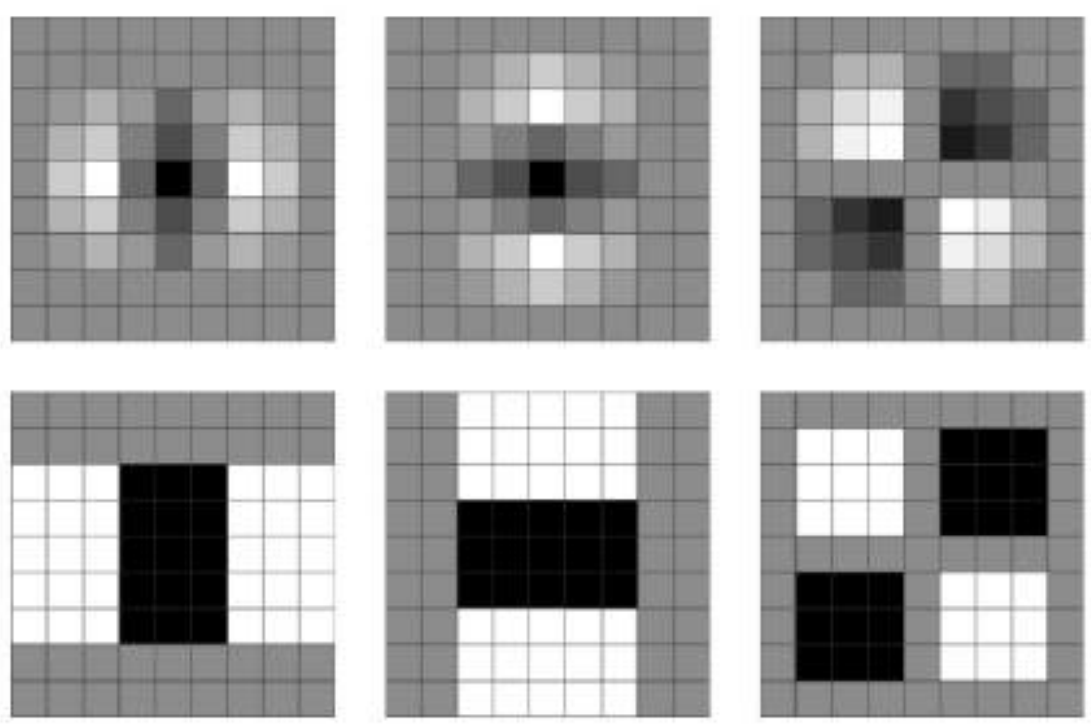

Figure 4.9: Laplacian of Gaussian approximations [40]. First row represents second order Gaussian derivatives in the $x$, $y$, and $xy$ directions, while second row shows corresponding weighted box filter approximations in the same directions.

Keypoint localization is performed both in scale and image space by applying kernels of increasing size to the original image, this is because the processing time of kernels used in SURF is size invariant. The responses of applied kernels are thresholded. Thus, increasing the threshold leaves only the strongest points while decreasing allows for many points to be detected. Next, a non maximum suppression in a $3 \times 3 \times 3$ neighborhood is performed where a pixel is selected as a keypoint if it is greater than the surrounding pixels on its interval and interval above and below. In the latter, each detected interest point is assigned a reproducible orientation to achieve a rotation invariance. Haar wavelet responses of size 4(σ) are computed for a number of pixels within a radius of 6(σ) as presented in Fig. 4.10. Haar wavelets are simple filters that can be employed to find gradients in the x and y direction as shown in Fig. 4.11(a). Then, a square window of a size 20(σ) is constructed around the interest point, furthermore the window is sub divided into 4 × 4 regular regions. Within each of these subregions Haar wavelets of size 2(σ) are computed for 25 regularly distributed sample points as shown in Fig. 4.11(b). The descriptor encompasses ($\sum d_x$,$\sum d_y$,$\sum |d_x|$,$\sum |d_y|$) where $d_x$ and $d_y$ are the $x$ and $y$ wavelet responses, thus the descriptor dimension will be 4×4×4=64 D. The final SURF descriptor is invariant to rotation, scale, brightness, and contrast after reduction to a unit length.

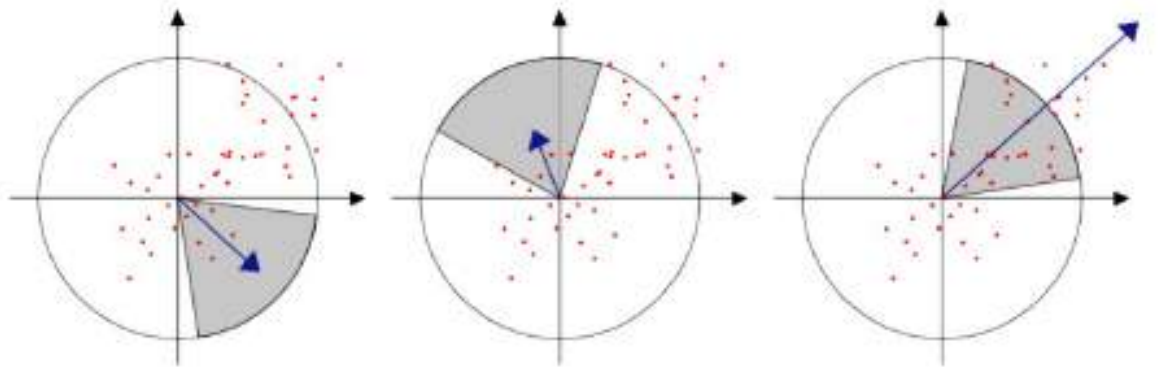

Figure 4.10: Orientation assignments [40]. When the window moves around the origin (60 degrees) the components of the responses are collected to yield the vectors shown in blue. The largest such vector determines the dominant orientation.

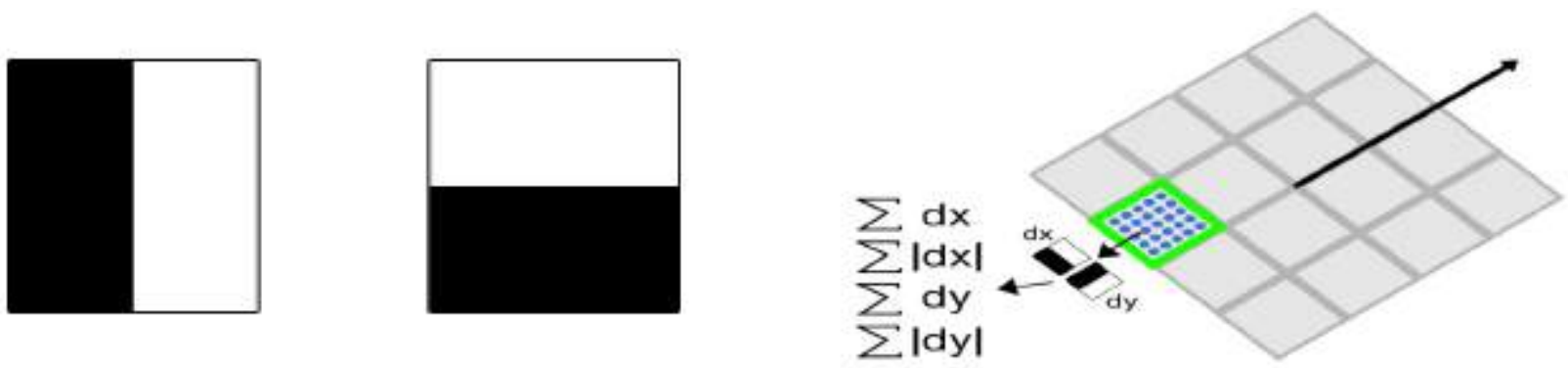

(a) (b)

Figure 4.11: Haar wavelets and descriptor components [40]. (a) The left filter computes the response in the $x$ direction and the right the $y$ direction. Weights are 1 for black regions and -1 for the white. (b) The green square encapsulates one of the 16 subregions and blue circles shows the sample points at which wavelet responses are computed.

### 4.2.2 Histogram of oriented gradients (HOG)

The histogram of oriented gradients (HOG) introduced by Dalal et al. [41] are feature descriptors used in computer vision and image processing for object recognition. The overall process can be summarized as discussed below:

**Gradient Computation**

Simple finite difference masks [-1,0,1] and its transpose work best with no Gaussian smoothing (σ=0).

**Weighted vote into spatial and orientation cells**

The image is decomposed into square cells of a given size. Then, a histogram of oriented gradients is computed in each cell with fixed number of predetermined bins. The orientation bins are spaced over [0 °, 180 °] (unsigned gradients). The gradient magnitudes of the pixels in the cell are used to vote into the orientation histogram. The magnitude and phase are computed respectively as follows:

$$\|\nabla F\| = \sqrt{\left(\frac{\partial f}{\partial y}\right)^2 + \left(\frac{\partial f}{\partial x}\right)^2} \tag{4.5}$$

$$\theta = \arctan\left(\frac{\partial f}{\partial y}/\frac{\partial f}{\partial x}\right) \tag{4.6}$$

**Contrast normalization over overlapping spatial blocks**

Cells are grouped into larger spatial blocks with 50% overlapping as shown in Fig. 4.12. Next, each block is $L_2$ normalized in order to provide better invariance to illumination, shadowing, and edge contrast. Each cell is shared between several blocks because of the overlapping. However, its normalization is block dependent and thus different. The cell appears many times in the final output vector along with different normalizations. This may look redundant but it improves the result.

**Collect HOG'S over detection window**

Each block consist of a certain number of cells (typically 2×2) and each cell contains $(n \times n)$ pixels with fixed number of bins for each cell (typically 9). Each cell yields a local 1-D histogram of gradients over all the pixels in the cell, for a total of 4×(number of orientations)=36.

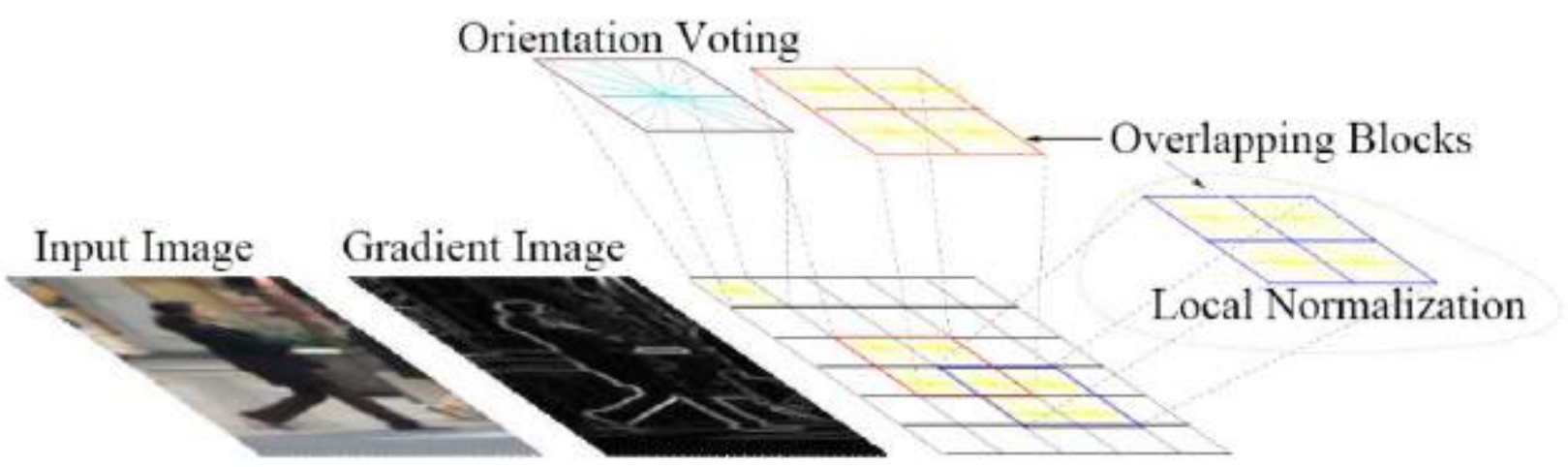

Figure 4.12: Histogram of oreinted gradients [http://www.vision.rwth-aachen.de].

The HOG presented by Felzenszwalb et al. [42] computes directed, undirected gradients, and a four dimensional texture energy feature. The directed gradient (sensitive gradient [0 °, 360 °]) is specified according to the contrast direction with a dimension of 2×(number of orientations). The undirected gradient (insensitive gradient [0 °, 180 °]) doesn't depend on the contrast direction with half the size. Here each cell within a block is individually $l_2$ normalized opposite to [41]. In total 4×(2+1)×(number of orientations) elements are yielded. However, the result is projected down to (2+1)×(number of orientations) by averaging corresponding histogram dimensions. The four dimension texture energy feature is obtained by $l_1$ normalizing each of the already $l_2$ normalized undirected histogram [43]. Final dimension becomes (2+1)×(number of orientations)+ additional four dimensions=31.

### 4.2.3 Local Binary Pattern (LBP)

The local binary pattern (LBP) is a simple very efficient texture descriptor that categorizes local image primitives such as curved edges, spots, and flat areas into a feature histogram. The original implementa-tion of the LBP operator [44, 45] at a given pixel is performed by thresholding the 3×3 neighborhood of every pixel with the center's pixel value. Assume $g_{\mathrm{c}}$ be the center pixel gray level and $g_{\mathrm{p}}(p = 0, 1, 2, ..., 7)$ be the gray level of each surrounding pixel. If $g_{\mathrm{i}}$ is smaller than $g_{\mathrm{c}}$, so the binary result of the pixel is set to 0 , otherwise to 1. All results are aggregated to a 8-bit binary value. The decimal value of the binary is the LBP feature. Fig. 4.13 shows an example of computation of LBP operator in a 3×3 neighborhood.

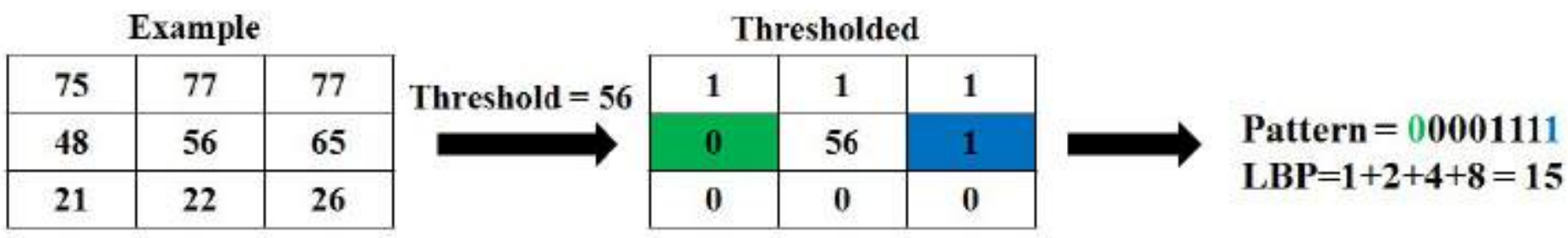

Figure 4.13: Illustration of the basic LBP operator.

The standard LBP operator is not very robust against local changes in the texture such as changes in viewpoint or illumination. As a result, local features computed in a 3 × 3 neighborhood cannot capture large scale structures that may be the only relevant texture features. In order to solve this problem, it is important to extend this idea to a multi resolution LBP operator [46]. The new idea is to use a set of points equally spaced on a circle centered at a pixel to be labeled. Thus, this permits any radius and number of sampling points. If a sampling point doesn't fall in the center of a pixel, bilinear interpolation is used. Fig. 4.14 shows an example for new LBP operator computation.

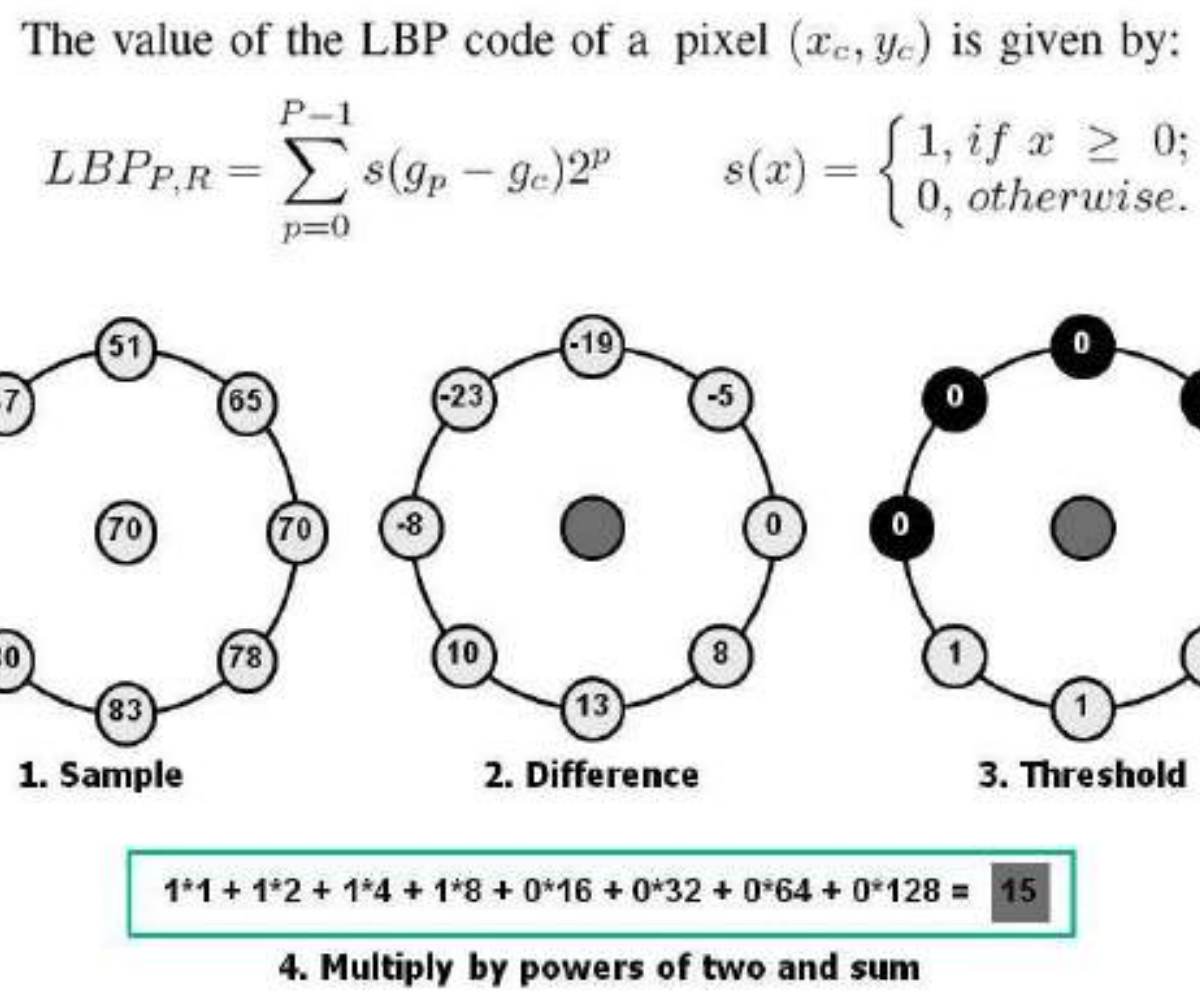

Figure 4.14: An example of new LBP operator computation. [http://www.scholarpedia.org/].

Another extension using uniform patterns is proposed, which can be applied in order to reduce the length of the feature vector while keeping its discriminative power. A LBP pattern is called uniform if the binary pattern encompasses maximum two bitwise transitions from $0 - 1$ or from $1 - 0$. For instance, the patterns (00000000) (no transitions), (01110000) (two transitions) are uniform, whereas (11001001) (4 transitions) and (01010010) (6 transitions) are not uniform. This constraint reduces the number of the LBP patterns from 256 to 58 for eight points [47].

## 4.3 Codebook Generation

K-means clustering algorithm is one of the most popular vector quantization methods. The purpose of the K-means is to partition N measurements into K clusters in which each observation belongs to the cluster with the closest mean. Assume the set of features extracted from the training set can be expressed as $\{x_1, x_2, \ldots, x_M\}$ where $x_m \in \mathbb{R}^{\mathrm{D}}$, the goal is to partition this feature set into K clusters $\{d_1, d_2, \ldots, d_K\}$ $d_k \in \mathbb{R}^{\mathrm{D}}$. Consider for each feature $x_m$, there is a corresponding set of binary indicator variables $r_{mk} \in \{0, 1\}$. If $x_m$ is appointed to cluster K, then $r_{mk} = 1$ and $r_{mj} = 0$ for $j \neq k$. The objective function can be defined by Eq. (4.7):

$$\min H(\{r_{mk}, d_k\}) = \sum_{i=1}^{M} \sum_{i=1}^{K} r_{mk} \|x_m - d_k\|^2 \tag{4.7}$$

The main idea is to find values for both $r_{mk}$ and $d_k$ to minimize the objective function $H$. Usually, this function is optimized in an iterative procedure where each iteration consists of two successive steps relating to successive optimization of $r_{mk}$ and $d_k$ [48]. Fig. 4.15 shows an example of visual word generation from a set of features using K-means.

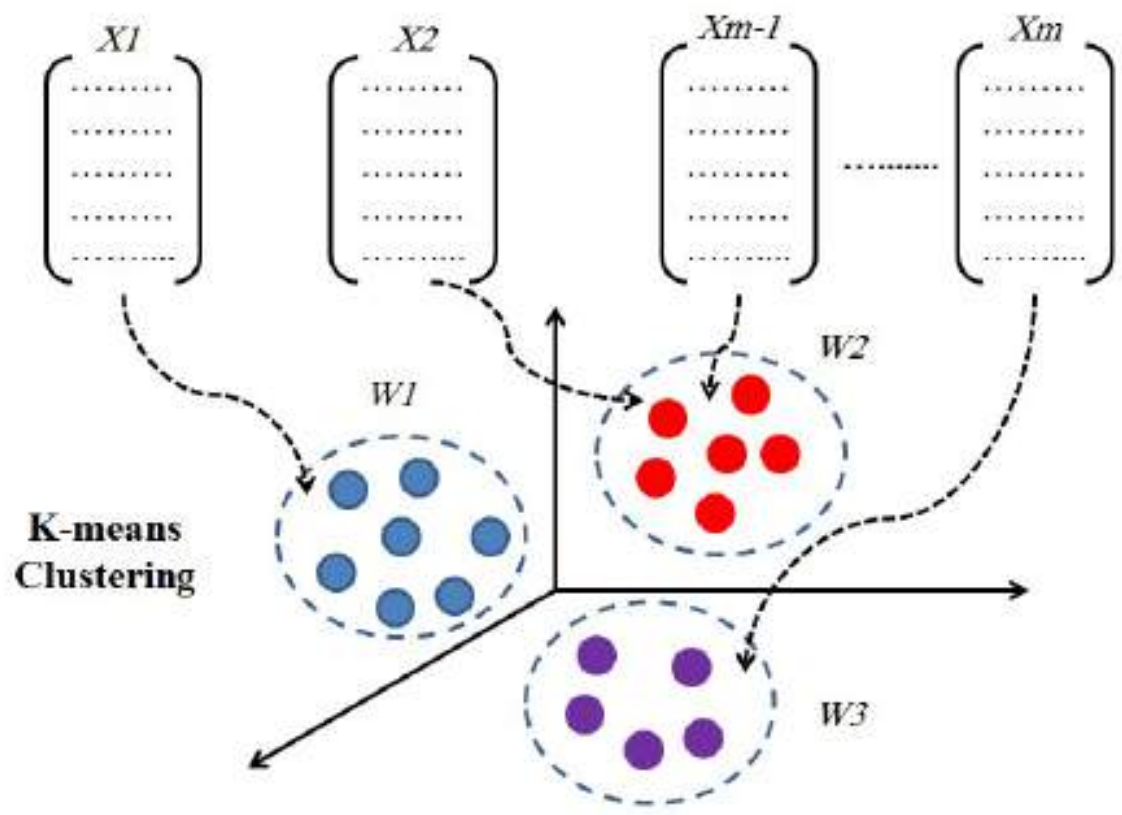

Figure 4.15: Codebook generation using k-means clustering algorithm. $\{x_1, x_2, \ldots, x_M\}$ are the feature sets and $\{w_1, w_2, \ldots, w_K\}$ represent the visual words.

Two criteria are introduced in order to construct the visual dictionary such as; a single based criterion and a multiple based criterion. In the former, a single dictionary $D_i$ is created from a pool of features, i.e. DSURF, SURF, HOG, or LBP. Then, for each image feature $x_m$find its corresponding visual word from every dictionary $D_i$. These visual words are combined into individual histograms $h_i$ for each dictionary (see Section 4.4) and the system performance is assessed accordingly. In the latter, similar steps are followed. However, the individual histograms are concatenated into a single one i.e. $h = [h_1, h_2, \ldots, h_N]$. These two methods are shown respectively in Fig. 4.16 and Fig. 4.17.

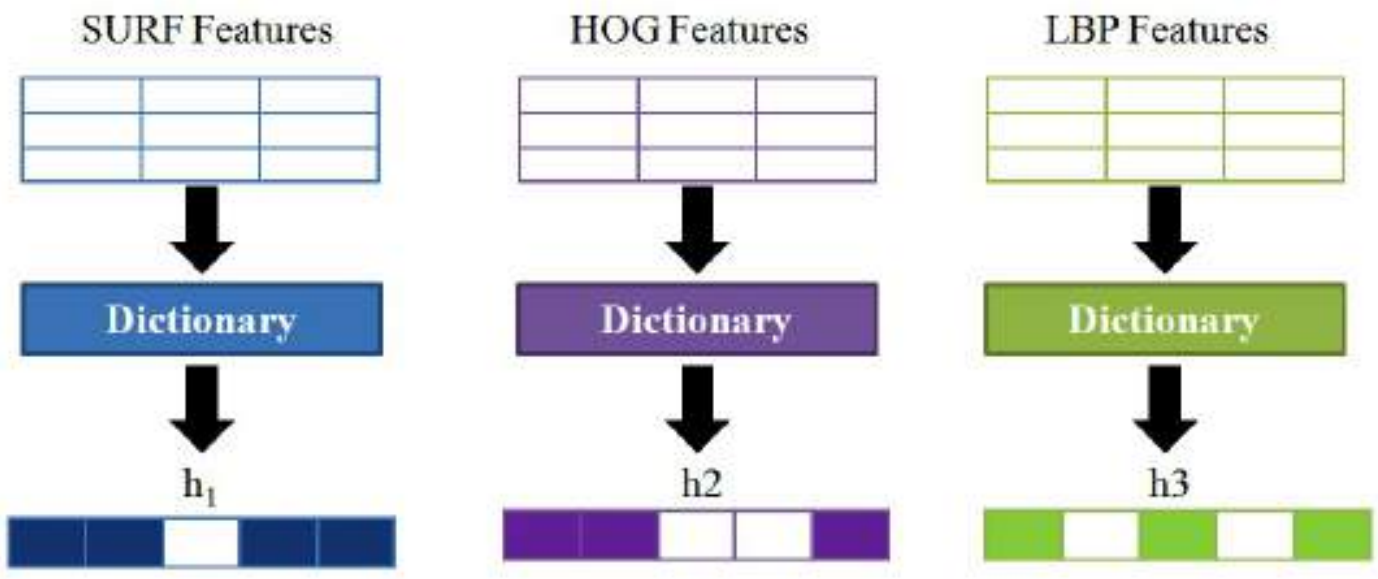

Figure 4.16: Single based dictionary example. The set of features represent a single image in the training dataset.

## 4.4 Feature Encoding and Pooling

For each individual image in the training dataset, consider $X$ be a set of D-dimensional descriptors such as $X = \{x_1, x_2, \ldots, x_N\} \in \mathbb{R}^{\text{D x N}}$. Given a visual dictionary (the one already computed in the previous step) with $K$ visual words i.e. $D = \{d_1, d_2, \ldots, d_K\} \in \mathbb{R}^{\text{D x K}}$. The purpose of the encoding step is to compute a code for input $x$ with $D$. Thus, each feature descriptor $x_n$ is allocated to the nearest visual word in the dictionary by satisfying $\arg\min_k \|x_n - d_k\|^2$. As a result a single vector $P$ containing the corresponding words of each feature descriptor $x_n$ is obtained; this is usually called hard assignment coding [48]. The pooling process is simply performed by counting the number of occurrences of each visual word in the resultant vector $P$, then $L_2$-normalize this vector as follows:

$$P = \frac{P}{\sqrt{(\sum_{k=1}^{K} p_k^2}} \tag{4.8}$$

To conclude, each image will be represented by a single normalized feature histogram corresponding to the number of occurrences of each visual word in that image. Subsequently, these histograms will fed into a linear support vector machine for final classification which will be explained in Section 4.5.

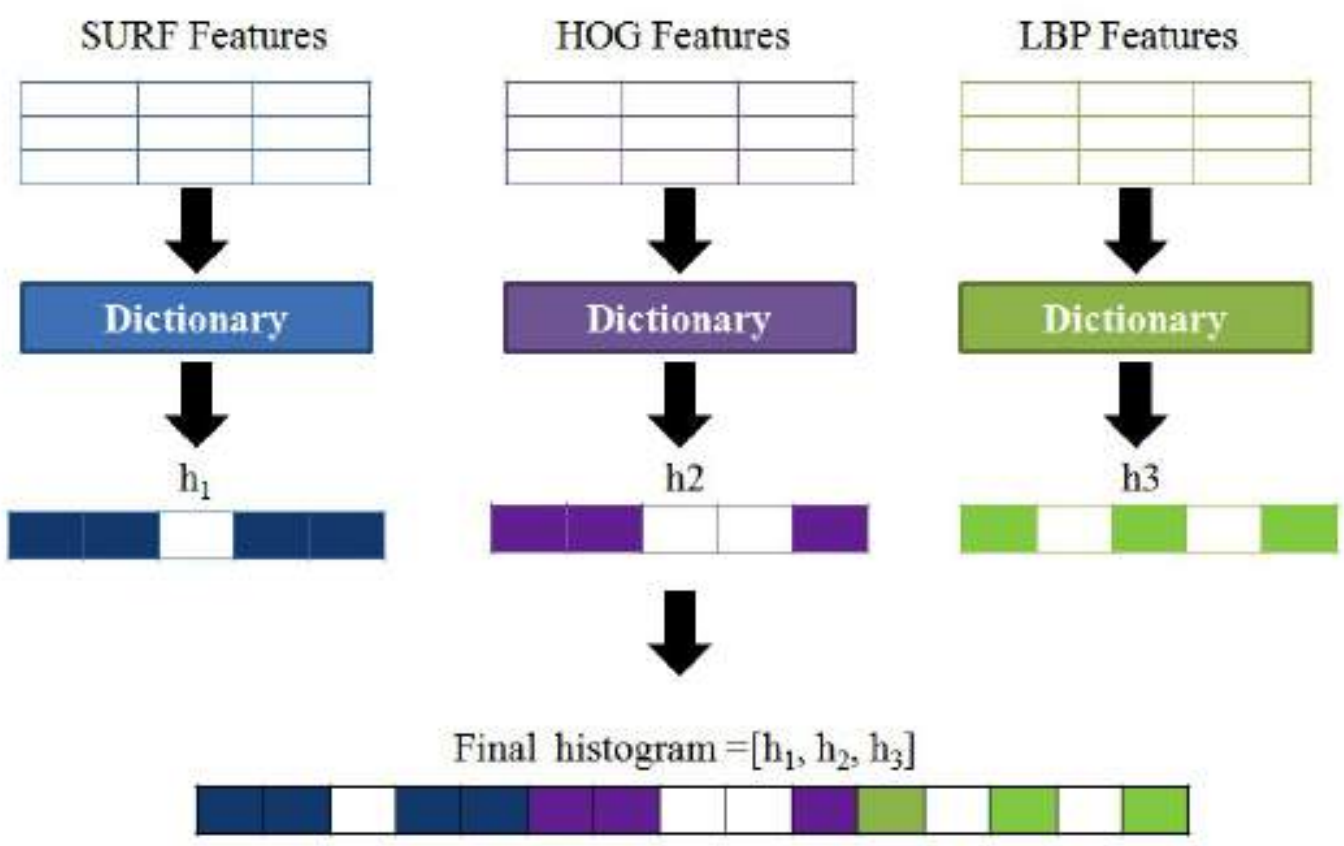

Figure 4.17: Multiple based dictionary example. The set of features represent a single image in the training dataset.

## 4.5 Classification

The problem of the classification has been performed using LIBSVM[9]. It is a popular open source machine learning library developed by National University of Taiwan and written in C++ [49]. Support vector machines (SVMs) are usually adapted for the purpose of data classification. Most probably, in any classification problem the data is separated into training and testing sets. Each example in the training set contains a class label and several features. Based on the training data, the objective of support vector machine is to produce a model which is able to estimate the target values of the test data given only the test data features. Assume a training set of instance pair of labels $(x_i, y_i)$, $i = 1, 2, \ldots, l$ where $x_i \in \mathbb{R}^{\mathrm{n}}$ and $y \in \{1, -1\}^l$ such that $y = +1$ for positive samples and $y = -1$ for negative samples, the SVM requires solution of the following lagrange optimization problem:

$$\begin{aligned} \min_{w,b,\xi} \quad & \frac{1}{2} w^T w + C \sum_{i=1}^{l} \xi_i \\ \text{subject to} \quad & y_i \left( w^T \phi(x) + b \right) \geq 1 - \xi_i \\ & \xi_i \geq 0 \end{aligned} \tag{4.9}$$

The training vectors $x_i$ are mapped into a higher dimensional space by the kernel function $\phi(x)$. The SVM finds a linear hyperplane which maximizes the margin $\frac{1}{2} w^T w$ in this higher dimensional space as we can see in Fig. 4.18. $C > 0$ is the penalty parameter of the error term. This parameter is referred to as *bestc* which should be tuned carefully during the training phase since it significantly affects the classifier performance. There are different kernel functions available i.e. linear, polynomial, radial basis function, and sigmoid. However, in our case we consider the linear one since it provides us with the best classification results. The linear kernel function is defined as $K(x_i, x_j) = x_i^T x_j$.

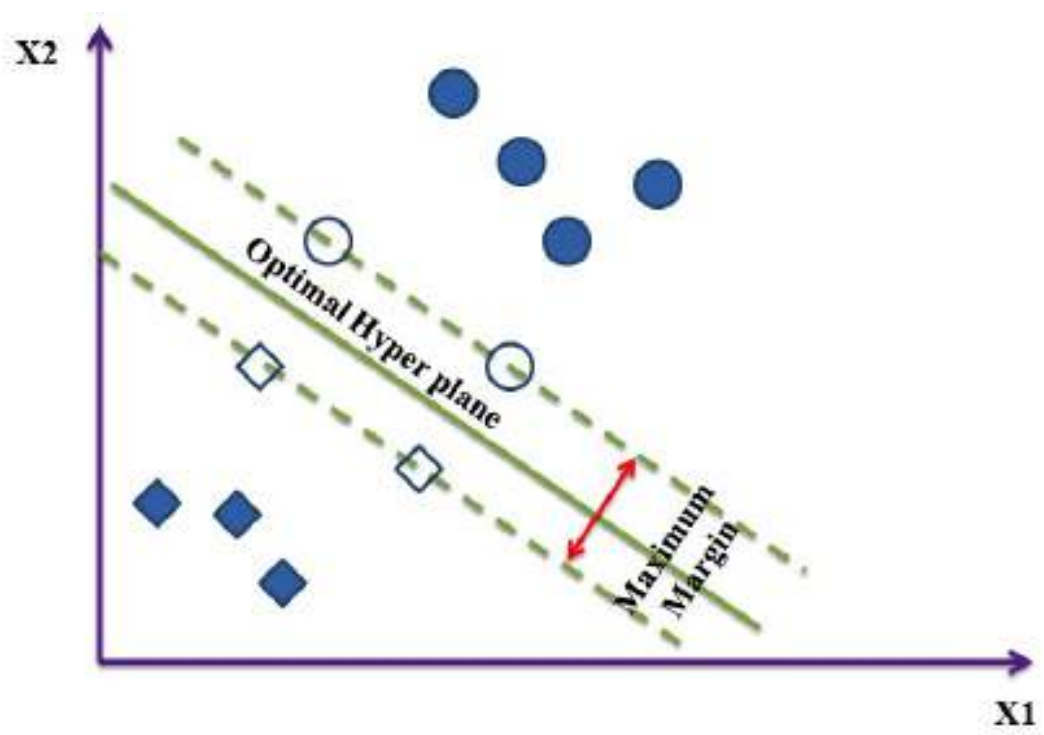

Figure 4.18: SVM optimal hyperplane for a set of 2D-points

It seems reasonable to conclude that, this chapter describes in detail the preprocessing steps as well as feature extraction techniques. Furthermore, it explains the pipeline of the bag of words approach besides a brief summary about the support vector machine. The next chapter will describe the different datasets being used, in addition to results and discussion.

[9] `http://www.csie.ntu.edu.tw/~cjlin/libsvm`

# Chapter 5

# Results and Discussion

In this research, we have used 430 images from six publicly available datasets such as; STARE[10],DRIVE[11], DRIDB[12], HEI-MED[13], MESSIDOR[14], and HRF[15], in addition to one private dataset obtained from the Oak Ridge National Laboratory, USA (ORNL). The description and distribution of these datasets are described in the following two sections.

## 5.1 Datasets Description

### 5.1.1 STARE

STARE (structural analysis of retina) dataset [52] consists of 400 PPM images, these images are digitized slides captured by a TopCon TRV-50 fundus camera with 35 degree field of view. Each slide was digitized to produce 605 $\times$ 700 pixel image with 24 bits per pixel. Annotations at image level are available. Fig. 5.1 shows an example of an image from the database.

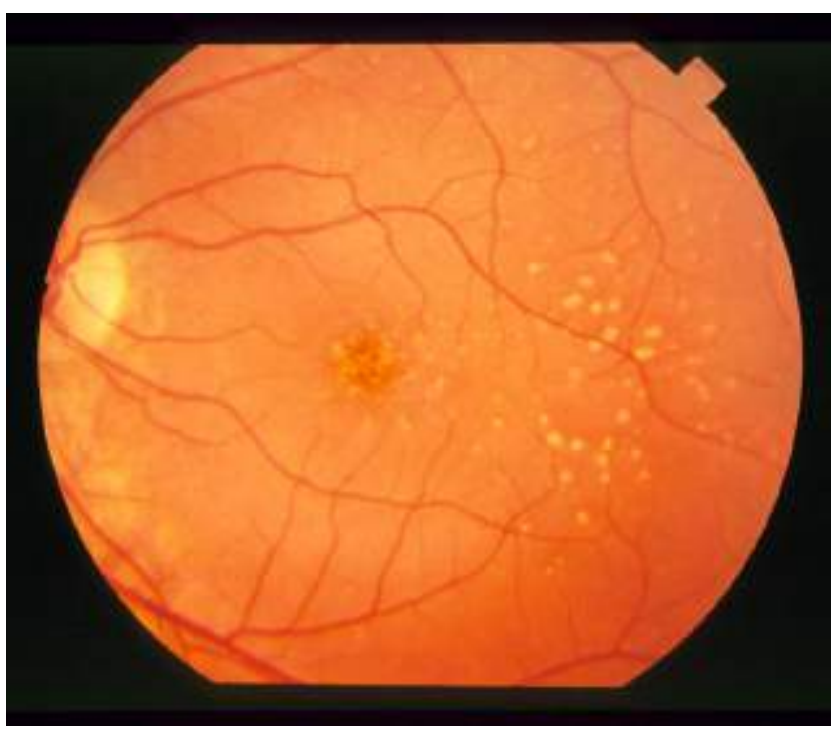

Figure 5.1: Drusen image from the STARE database.

[10]see (http://www.ces.clemson.edu/~ahoover/stare/)
[11]see (http://www.isi.uu.nl/Research/Databases/DRIVE/)
[12]see (http://www.fer.unizg.hr/ipg/resources/image_database)
[13]see (http://vibot.u-bourgogne.fr/luca/heimed.php)
[14]kindly provided by the Messidor program partners (see http://messidor.crihan.fr)
[15]see (http://www5.cs.fau.de/research/data/fundus-images/)

### 5.1.2 DRIVE

DRIVE (digital retinal images for vessel segmentation) dataset [53] contains 40 TIFF images and their groundtruth (vessel segmentation). Out of the 40 images in the database, 7 contains diseases like exudates, hemorrhages and pigment epithelium changes. All images are digitized using a canon CR5 non-mydriatic 3CCD camera with 45 degree field of view. Every image is captured using 24 bits per pixel at image size of 584 × 565 pixels. Fig. 5.2 presents an example of such image.

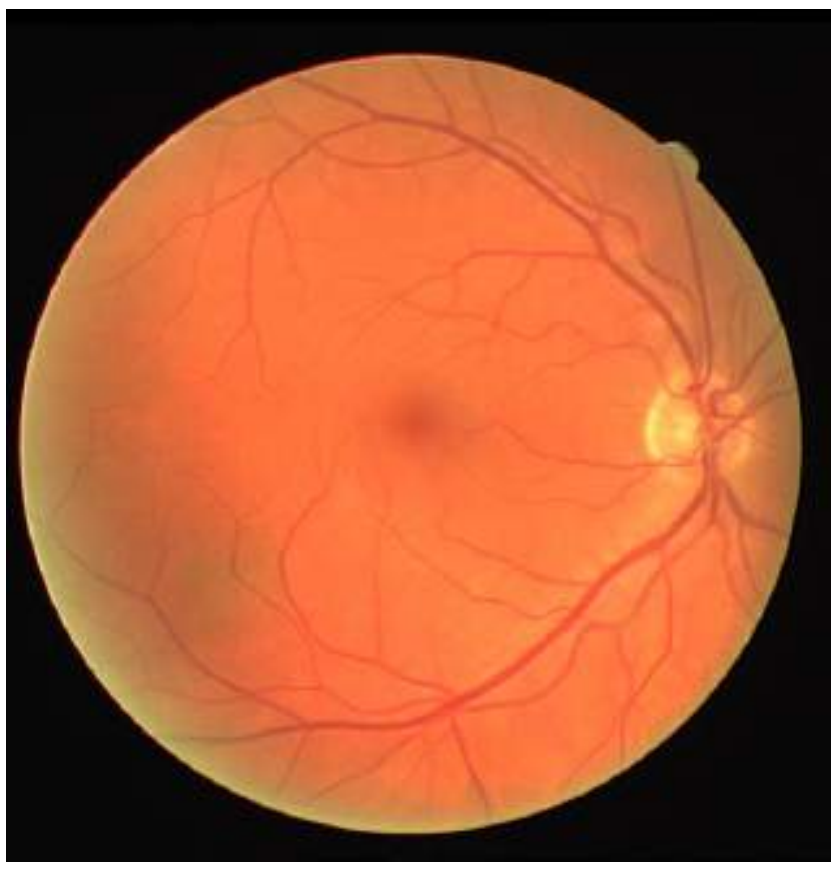

Figure 5.2: Normal image from the DRIVE database.

### 5.1.3 DRIDB

DRIDB (diabetic retinopathy image database) dataset [54] encompasses 50 BMP images and their groundtruth (microaneurysms, hemorrhages, hard exudates, soft exudates, blood vessels, optic disk, and macula). The images are captured at a resolution of 720 × 576 in RGB color with 8 bits per color plan with a ZEISS VISUCAM 200 fundus camera at 45 degree field of view. Fig. 5.3 shows one such image.

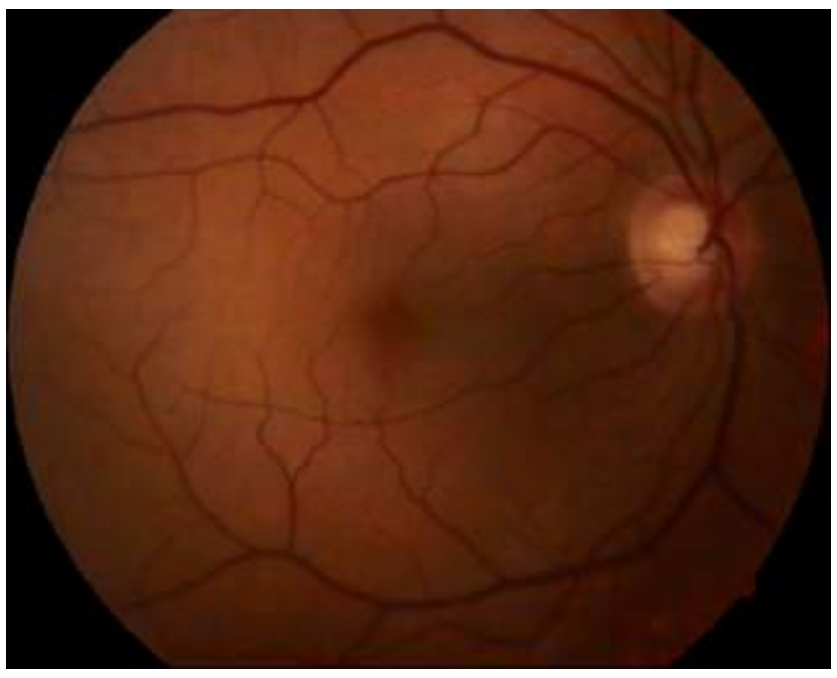

Figure 5.3: Normal image from the DRIDB database.

### 5.1.4 HEI-MED

HEI-MED (Hamilton eye institute macular edema) dataset [29] is a combination of 169 JPEG images (high quality fundus images from different ethnic backgrounds). Furthermore, a manual groundtruth lesion map and other meta-data are available. It is mainly used for exudates and diabetic macular edema detection. The images are acquired by ZEISS VISUCAM PRO fundus camera at a resolution of 2196 × 1958 and with a field of view 45 degree. Two examples are shown in Fig. 5.4.

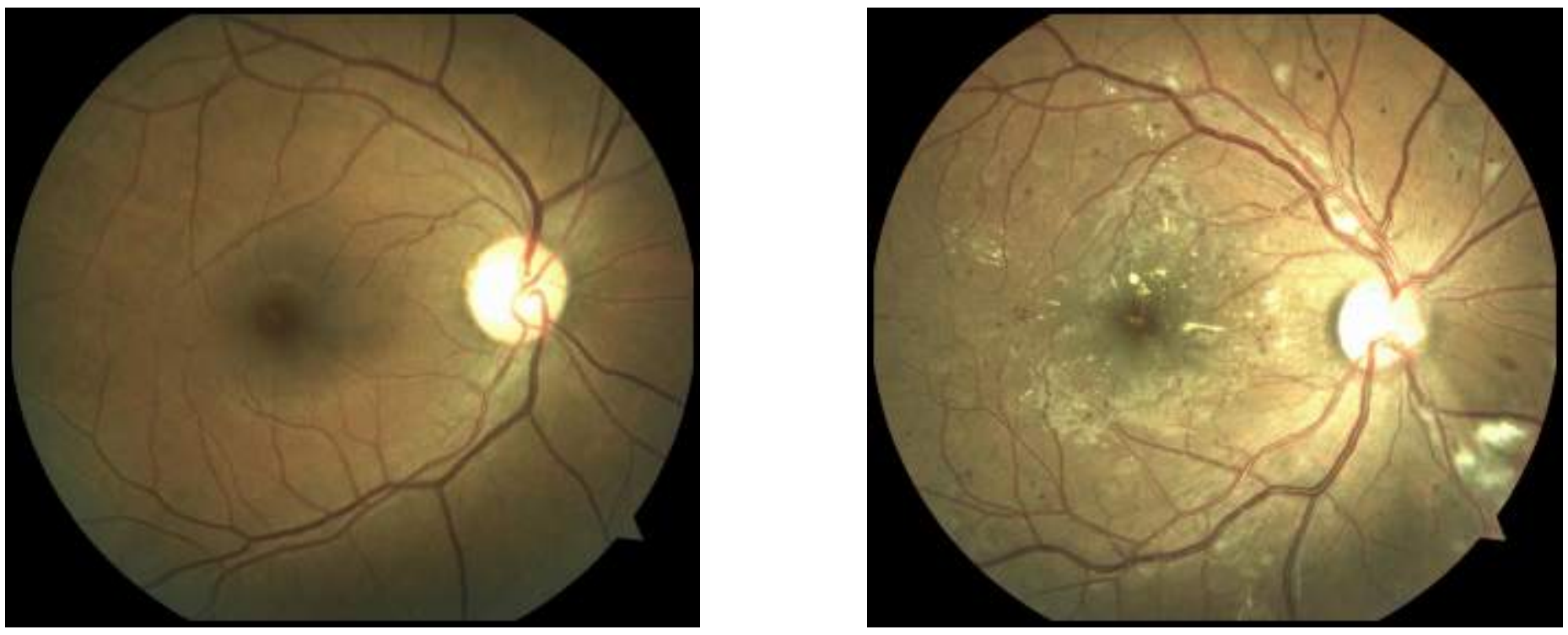

(a) (b)

Figure 5.4: Two examples of normal and exudate images from HEI-MED database.

### 5.1.5 MESSIDOR

MESSIDOR (method to evaluate segmentation and indexing techniques in the field of retinal ophthalmology) dataset [55] combines 1200 TIFF images. These images are acquired using a color video 3CDD camera on a TopCon TRC NW6 non-mydriatic retinopathy with 45 degree field of view with 8 bits per color plane at 1440 × 960, 2240 × 1488 or 2304 × 1536 pixels. 800 images were acquired with pupil dilation and 400 without dilation. The images are taken at three different clinical sites. Grading for diabetic retinopathy and the risk of macular edema in each image are provided. For the best of our knowledge, it is the largest available images on the internet. An example of such images is shown in Fig. 5.5

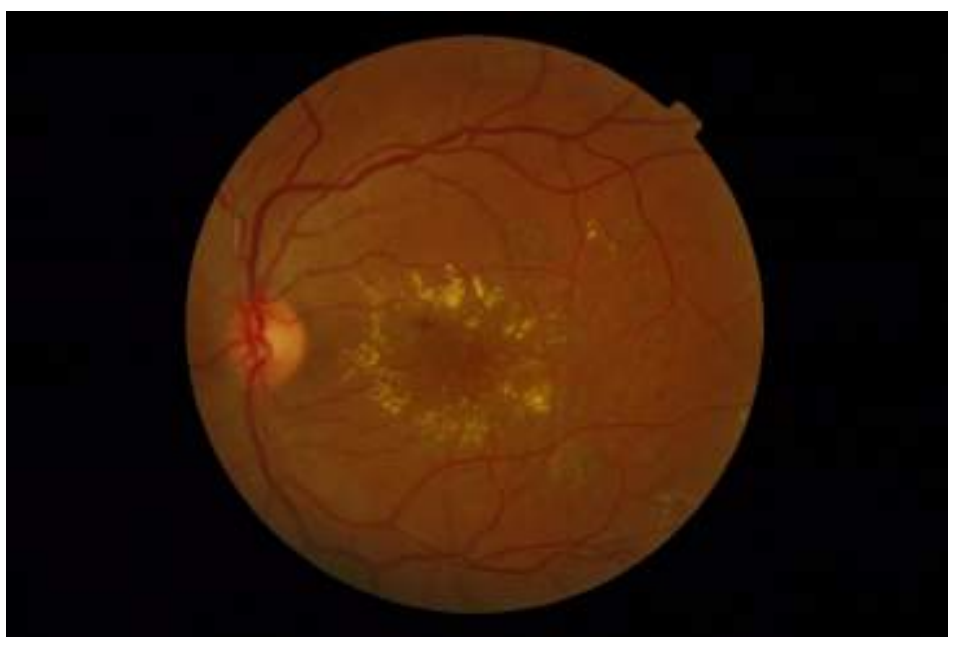

Figure 5.5: Exudate image from the MESSIDOR database.

### 5.1.6 HRF

HRF (high resolution fundus) dataset [56] comprises 15 images of healthy patients, 15 images of patients with diabetic retinopathy, and 15 images of glaucomatous patients. The images are acquired using a canon CR-1 fundus camera at a resolution of 2336 × 3504 with a field of view 45 degree and different acquisition setting. Groundtruth for vessel segmentation is available. Fig. 5.6 shows an example.

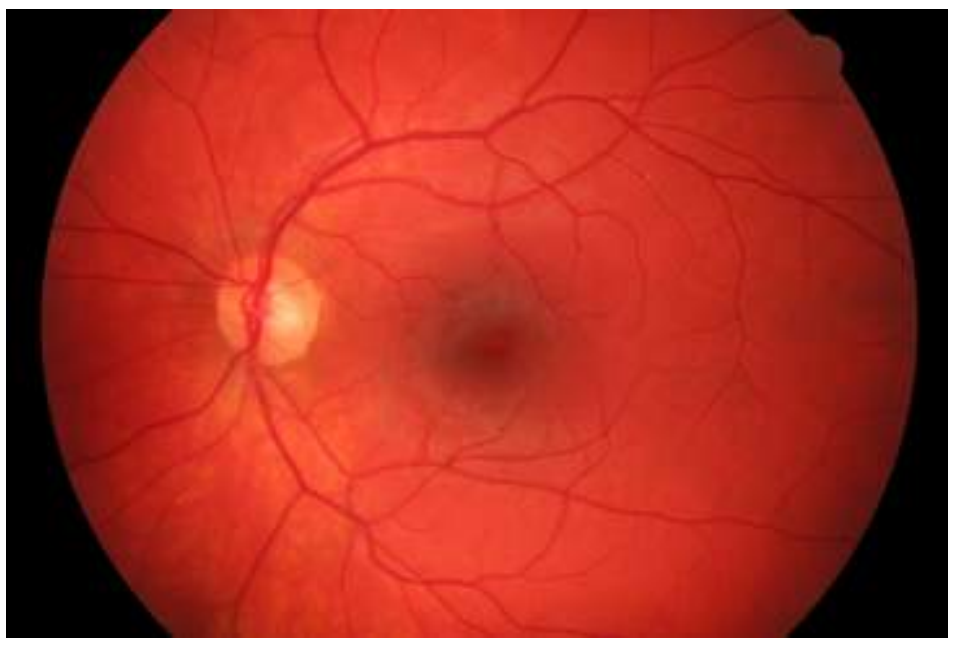

Figure 5.6: Normal image from the HRF database.

### 5.1.7 ORNL

ORNL (Oak Ridge National Laboratory, USA) dataset is captured at a resolution of 1024 × 1360 and 1958 × 2196 with 45 degree field of view. The dataset consists of 61 drusen only images, 36 normal images, and 20 exudates only images. No groundtruth is available. Two examples are presented in Fig. 5.7.

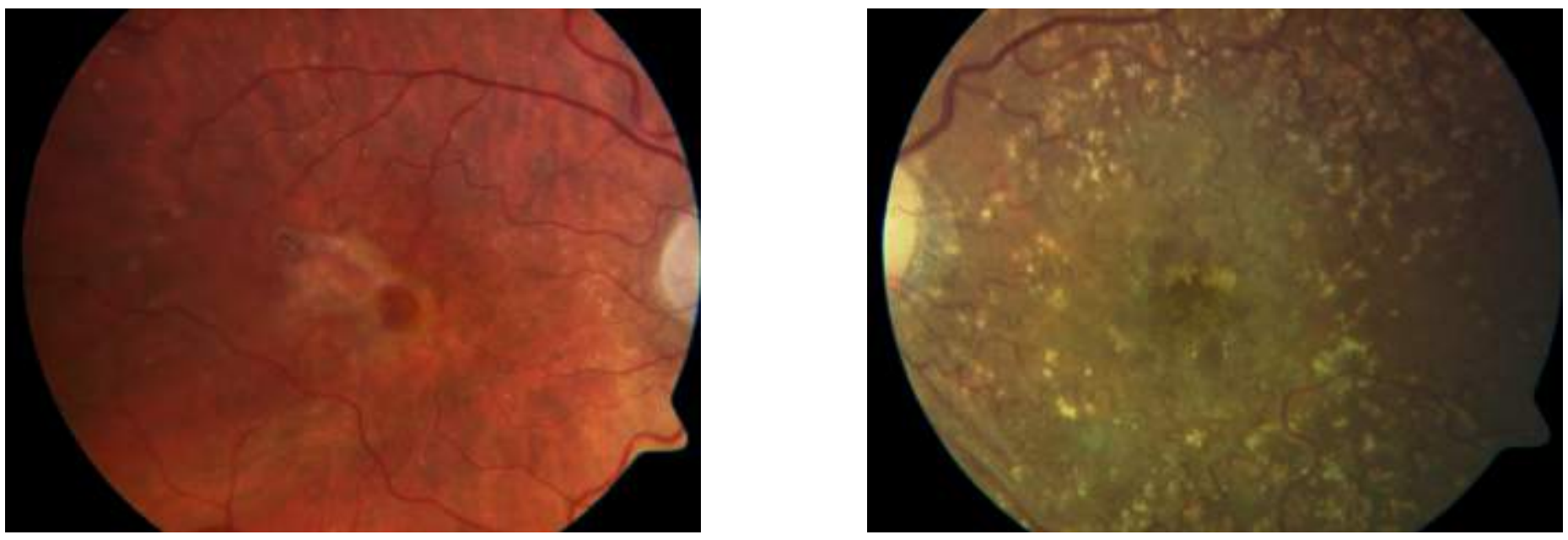

(a) (b)

Figure 5.7: Two drusen examples of ORNL database.

## 5.2 Datasets Distribution

In this study, we have used 81 normal images, 85 drusen images, and 264 exudate images obtained from (ORNL, HRF, DRIDB, and DRIVE), (ORNL and STARE), and (ORNL, HEI-MED, and MESSIDOR) respectively. The images are divided into two sets; Set A and Set B. Set A contains 220 images acquired from ORNL, HEI-MED, HRF, DRIVE, DRIDB, and only one clinical site of MESSIDOR named MES1 whereas Set B constitutes 210 images obtained from ORNL, HEI-MED, HRF, DRIVE, DRIDB, and two clinical sites of MESSIDOR named MES2 and MES3.

The idea is to use Set A as a training set, then measure the system performance based on Set B and vice-versa. In this way, we can assess how well the system behaves when the test set contains different images than the ones included in the training set. This is usually called cross dataset testing. That means the proposed system (selected features, dictionary: single or multiple, and number of visual words) should be discriminating enough to classify the date present in the Set A based on Set B, also the data present in Set B based on Set A.

Table 5.1 shows Set A and Set B image distribution.

| | **SETA** | | |
|---|---|---|---|
| | **Normal** | **Drusen** | **Exudates** |
| **ORNL** | 18 | 30 | 10 |
| **HEI-MED** | ... | ... | 13 |
| **STARE** | ... | 12 | ... |
| **HRF** | 7 | ... | ... |
| **DRIDB** | 5 | ... | ... |
| **DRIVE** | 10 | ... | ... |
| **MSE1** | ... | ... | 115 |
| **MSE2** | ... | ... | ... |
| **MSE3** | ... | ... | ... |
| **Number of images** | 40 | 42 | 138 |
| | **SETB** | | |
| | **Normal** | **Drusen** | **Exudates** |
| **ORNL** | 18 | 31 | 10 |
| **HEI-MED** | ... | ... | 13 |
| **STARE** | ... | 12 | ... |
| **HRF** | 8 | ... | ... |
| **DRIDB** | 5 | ... | ... |
| **DRIVE** | 10 | ... | ... |
| **MSE1** | ... | ... | ... |
| **MSE2** | ... | ... | 63 |
| **MSE3** | ... | ... | 40 |
| **Number of images** | 41 | 43 | 126 |

Table 5.1: Data distribution of Set A and Set B. MES1: MESSIDOR site 1, MES2: MESSIDOR site 2, and MES3: MESSIDOR site 3.

## 5.3 Experimental Results

As discussed in Section 4.3, there exist two different methods to construct the visual dictionary: single based method and multiple based method. The goal of the single based method is to build a single dictionary using a set of features such as DSURF, SURF, HOG, or LBP whereas the multiple based method is based on constructing multiple dictionaries from all set of features, then combining all histograms into one histogram.

The system performance is assessed using the accuracy measurement which is computed as follows:

$$\text{Accuracy} = \frac{\text{Total number of correctly classified images}}{\text{Total number of images}}\ \% \tag{5.1}$$

or

$$\text{Accuracy} = \frac{T_N + T_P}{T_N + F_P + F_N + T_P}\ \% \tag{5.2}$$

where

- $T_P$ (true positive): Abnormal images correctly diagnosed as abnormal
- $F_P$ (false positive): Normal images incorrectly diagnosed as abnormal
- $T_N$ (true negative): Normal images correctly diagnosed as normal
- $F_N$ (false negative): Abnormal images incorrectly diagnosed as normal

The classifier's parameter i.e. the value of $C$ which is refered to as *bestc* is computed by carrying out a class classification with 10 fold cross validation. The idea is to break the training data into 10 sets of a size $\frac{\text{n}}{10}$, train on 9 datasets and test on 1. Then, repeat 10 times and take the mean accuracy. Since K-means clustering algorithm (hard assignment) is employed, different values of K are used such as K $= [10, 20, 30, 40, 50, 60, 70, 80, 90, 100]$ in order to achieve satisfactory classification results. Two experiments are performed. The first experiment is to use Set B as a training set and Set A as a test set, while the second experiment is to use Set A as a training set and Set B as a test set.

### 5.3.1 Performance using Set B as a training set and Set A as a test set

Regarding the single based dictionary, the highest accuracy 98.63% is obtained using DSURF descriptors at K=70, subsequently HOG, SURF, and LBP achieve accuracies of 97.27% at K=100, 85.91% at K=90, and 91.36% at K=80 respectively as shown in Table 5.2. Neither SURF nor LBP descriptors provide satisfactory results as expected. On the contrary, HOG gives approximately similar results to DSURF 97.27% at K=100. Since there is no preprocessing step to remove the optic disk, it might be confusing for the SURF or LBP descriptors to discriminate between normal and exudate images as the intensity characteristics of the optic disk are very similar to the exudate lesions.

Although DSURF descriptors achieve the best accuracy, there is still a problem because an exudate image is misclassified as normal which in fact can be considered as a disadvantage to the system. If a diabetic patient is misdiagnosed as normal (false negative), that means we are in troubles. However, if a normal patient is misdiagnosed as diabetic (false positive) he or she may repeat the test to make sure a bout the disease. False negative is more pernicious than false positive. Fig. 5.8 shows the confusion matrix of DSURF descriptors at K=70. All non diagonal elements on the confusion matrix represent misclassified data whereas diagonal elements represent correctly classified data. Thus, a good classifier will yield a confusion matrix with more diagonal elements.

The accuracy of the above example is computed as follows:

| Training: Set B, Testing: Set A | | | | | |
|---|---|---|---|---|---|
| | Single Dictionary | | | | Multiple Dictionaries |
| **K** | **DenseSURF** | **SURF** | **HOG** | **LBP** | **DSURF, HOG, LBP** |
| **10** | 75.4545 | 78.1818 | 90.4545 | 80 | 91.3636 |
| **20** | 88.1818 | 83.1818 | 93.6364 | 85.4545 | 96.3636 |
| **30** | 93.1818 | 84.0909 | 93.1818 | 88.6364 | 95.4545 |
| **40** | 95 | 83.6364 | 93.6364 | 88.6364 | 98.6364 |
| **50** | 97.7273 | 83.1818 | 95.9091 | 89.0909 | 97.7273 |
| **60** | 95.9091 | 83.6364 | 92.2727 | 88.6364 | 97.7273 |
| **70** | 98.6364 | 84.5455 | 96.3636 | 90 | 97.7273 |
| **80** | 97.7273 | 85 | 93.1818 | 91.3636 | 99.0909 |
| **90** | 97.7273 | 85.9091 | 95.9091 | 90.9091 | 98.6364 |
| **100** | 98.1818 | 85.4545 | 97.2727 | 89.5455 | 99.5455 |
| **Max** | 98.6364 | 85.9091 | 97.2727 | 91.3636 | 99.5455 |

Table 5.2: Test Set A versus Set B using different number of visual words.

| | | Predicted | | |
|---|---|---|---|---|
| | | **Normal** | **Drusen** | **Exudates** |
| **Actual** | **Normal** | **40** | **0** | **0** |
| | **Drusen** | **0** | **41** | **1** |
| | **Exudates** | **1** | **1** | **136** |

Figure 5.8: Confusion matrix of DSURF descriptors at K=70 (test Set A Vs. Set B).

$$\text{Accuracy} = \frac{40 + 41 + 136}{40 + 41 + 136 + 3} = 98.63\% \qquad (5.3)$$

On the other hand, multiple based dictionary approach overcomes the single based dictionary. At K=100, an accuracy of 99.54% is obtained. This is accomplished with only one false negative sample i.e. one exudate image is misidentified as normal as shown in Fig. 5.9. However, there are no any false positive cases like DSURF descriptors.

| | | Predicted | | |
|---|---|---|---|---|
| | | **Normal** | **Drusen** | **Exudates** |
| **Actual** | **Normal** | **42** | **0** | **0** |
| | **Drusen** | **0** | **42** | **0** |
| | **Exudates** | **1** | **0** | **137** |

Figure 5.9: Confusion matrix of multiple based dictionary at K=100 (test Set A Vs. Set B).

In fact for all values of K, multiple based dictionary approach achieves higher results than the single based method, except at K=70 DSURF descriptors result 98.63% is slightly better than multiple based 97.72%. Fig. 5.10 shows the resultant accuracy for all descriptors versus different values of visual words.

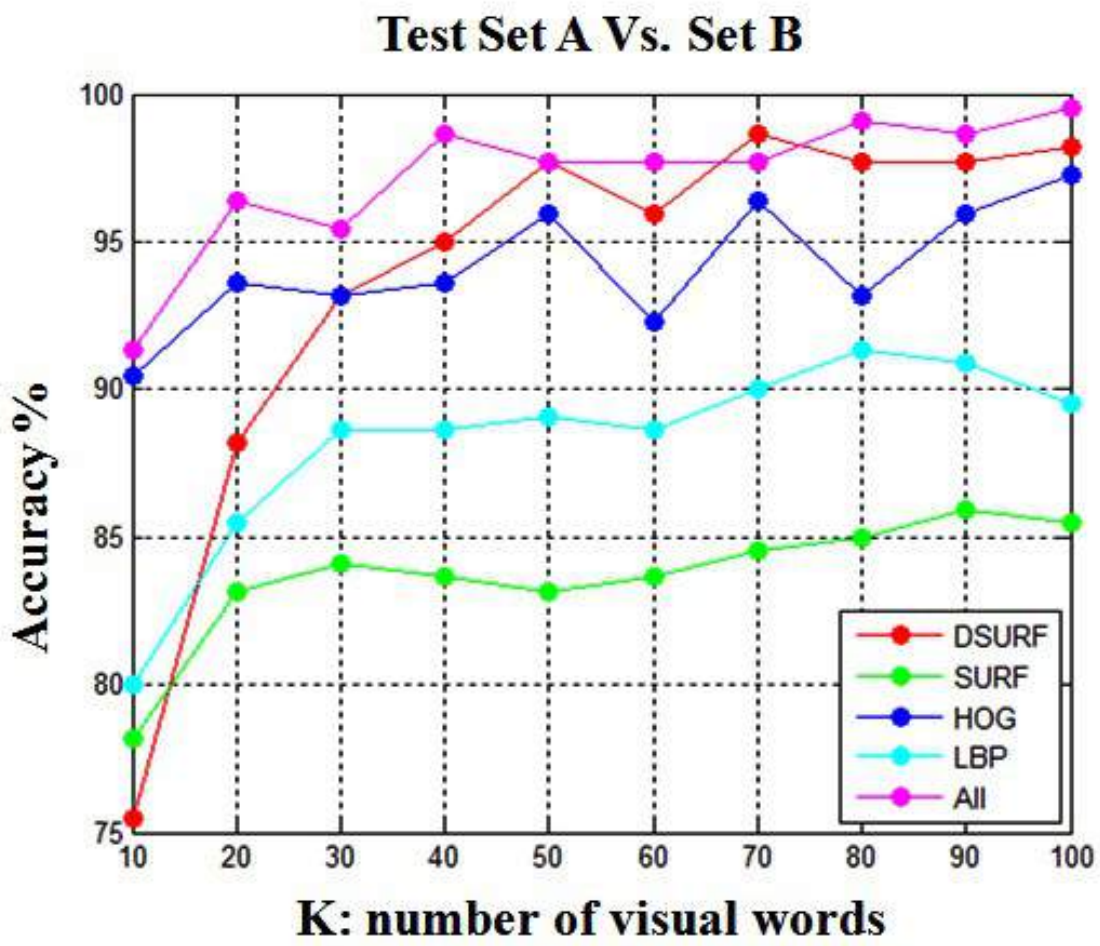

Figure 5.10: Accuracy Vs. visual words K for a single and multiple based dictionaries (test Set A Vs. Set B). All: multiple based dictionaries approach.

### 5.3.2 Performance using Set A as a training set and Set B as a test set

With respect to the single based dictionary, the HOG descriptors achieve the highest accuracy 97.14% at K=100, after that DSURF, SURF, and HOG achieve accuracies of 90.95% at K=50, 85.23% at K=80, and 84.76% at K=70 respectively as shown in Table 5.3.

| Training: Set A, Testing: Set B | | | | | |
|---|---|---|---|---|---|
| | Single Dictionary | | | | Multiple Dictionaries |
| **K** | **DenseSURF** | **SURF** | **HOG** | **LBP** | **DSURF, HOG, LBP** |
| **10** | 64.7619 | 73.3333 | 90.4762 | 78.5714 | 90.4762 |
| **20** | 77.1429 | 79.0476 | 92.8571 | 83.8095 | 95.7143 |
| **30** | 80.9524 | 80.4762 | 92.381 | 54.2857 | 98.0952 |
| **40** | 82.8571 | 80 | 95.7143 | 54.2857 | 98.0952 |
| **50** | 90.9524 | 81.4286 | 92.8571 | 77.1429 | 96.6667 |
| **60** | 89.0476 | 79.5238 | 94.7619 | 70.9524 | 97.619 |
| **70** | 86.6667 | 84.2857 | 96.1905 | 84.7619 | 98.0952 |
| **80** | 88.5714 | 85.2381 | 92.381 | 70.9524 | 99.5238 |
| **90** | 90.4762 | 83.3333 | 95.2381 | 77.1429 | 98.5714 |
| **100** | 90.4762 | 81.4286 | 97.1429 | 72.381 | 100 |
| **Max** | **90.9524** | **85.2381** | **97.1429** | **84.7619** | **100** |

Table 5.3: Test Set B versus Set A using different number of visual words.

The confusion matrix of the HOG descriptors shown in Fig. 5.11 indicates that two drusen images beside one exudate image are misidentified as normal, which remains a drawback to the system as we discussed before. The DSURF descriptors yield lower results 90.95% than the one obtained in the previous experiment. SURF and LBP descriptors attain relatively similar results as before. We can notice that the DSURF descriptors don't attain similar performances in both experiments owing to the sharp decrease in accuracy from 98.63% to 90.95%. However, the HOG descriptors achieve satisfactory results with the two experiments which implies the discriminative power of these descriptors.

| | | Predicted | | |
|---|---|---|---|---|
| | | Normal | Drusen | Exudates |
| Actual | Normal | 41 | 1 | 0 |
| | Drusen | 2 | 39 | 1 |
| | Exudates | 1 | 1 | 124 |

Figure 5.11: Confusion matrix of HOG descriptors at K=100 (test Set B Vs. Set A).

| | | Predicted | | |
|---|---|---|---|---|
| | | Normal | Drusen | Exudates |
| Actual | Normal | 42 | 0 | 0 |
| | Drusen | 0 | 42 | 0 |
| | Exudates | 0 | 0 | 126 |

Figure 5.12: Confusion matrix of multiple based dictionary K=100 (test Set B Vs. Set A).

Once more, the multiple based dictionary approach overcomes the single based dictionary since at K=100 a 100% accuracy is achieved as presented in the confusion matrix of Fig. 5.12, the false positive and false negative are zeros. Furthermore, for all visual words, it achieve higher results than the single based method as shown in Fig. 5.13. So far, we can conclude that the multiple based approach achieves significant results in both conditions, which indicates the importance of integrating several descriptors in the task of diabetic retinopathy diagnosis. As mentioned in Section 5.2 the proposed approach should be able to discriminate the data present in Set A based on Set B and vice-versa, the multiple based approach managed to accomplish this task with satisfying results such as 99.54%, 100% for the first and second experiment respectively and a mean accuracy of 99.77%.

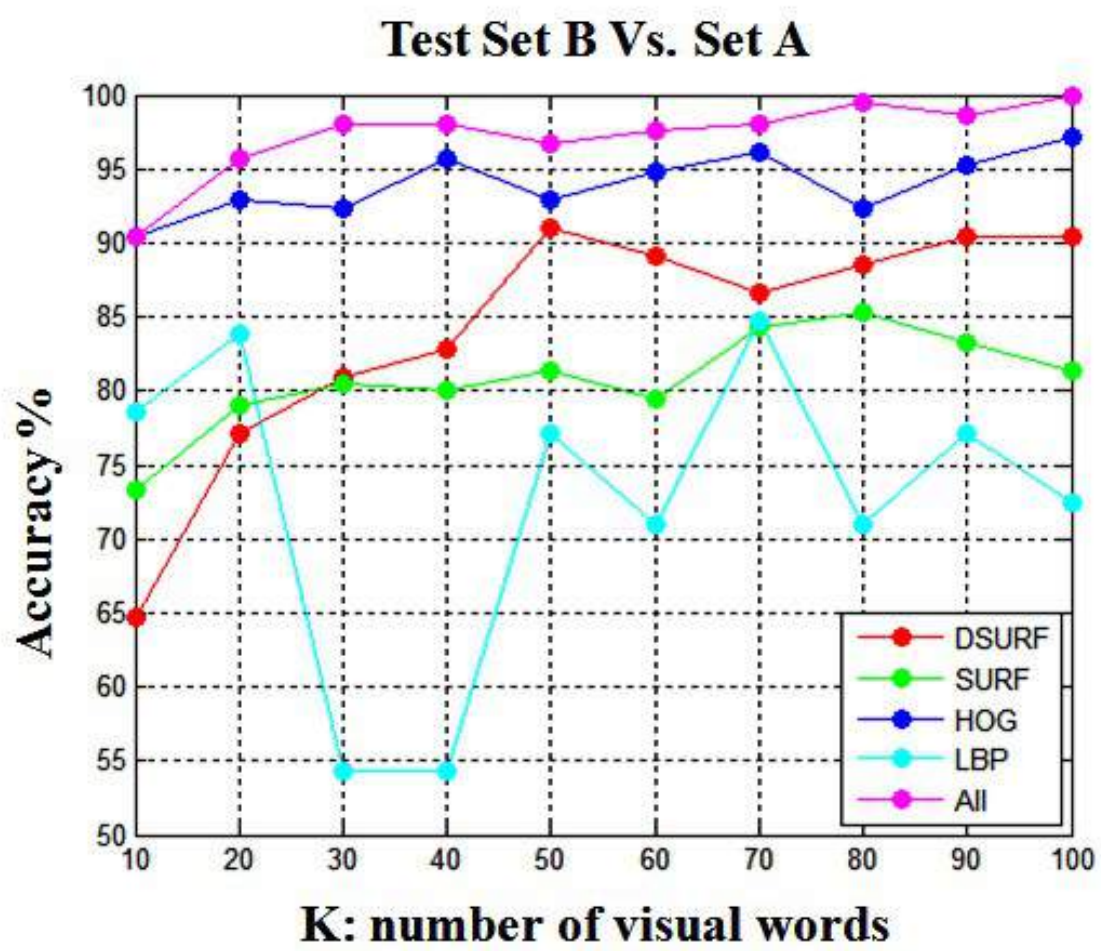

Figure 5.13: Accuracy Vs. visual words K for a single and multiple based dictionaries (test Set B Vs. Set A). All: multiple based dictionaries approach.

# Chapter 6

# Conclusion and Future Work

In this last chapter we present conclusions about our work. Moreover, we describe possible research directions to continue the work described in this thesis.

## 6.1 Conclusion

In this thesis, a bag of words approach was employed in order to discriminate between normal fundus images and abnormal fundus images with bright lesions, specifically drusen and exudates. We have proposed to use a single based and multiple based dictionary. In the first, a single dictionary is constructed from DSURF, SURF, HOG, or LBP descriptors, after that a histogram of word occurrences is generated for each image and the system performance is assessed accordingly. In the second, the image gets a histogram from each dictionary which are horizontally concatenated to form a single histogram, where each feature gets $N$ entries in the histogram, one from each dictionary.

The two schemes are evaluated on six publicly available datasets, namely STARE, DRIVE, DRIDB, HEI-MED, MESSIDOR, and HRF beside one local dataset obtained from the Oak Ridge National Laboratory, USA. The datasets have been divided into two sets: Set A and Set B. Set A contains data from the first clinical site of MESSIDOR whereas Set B comprises data from the second and third clinical sites of MESSIDOR. The goal was to evaluate Set A bases on Set B and vice-versa. In such way, we will be able to test how well the system generalize when the test set contains different data than the ones included in the training set.

We achieved a mean accuracy of 97.2% with respect to the single based dictionary, while our best accuracy is obtained using the multiple based dictionary with a mean accuracy of 99.77% which reflects the discriminative power of this approach.

To conclude, the bag of words approach can play a significant role in the classification of normal fundus images and abnormal fundus images with bright lesions, also it helps physicians in the early diagnosis of diabetic retinopathy as exudates might be the only sign of diabetic retinopathy as we discussed in Chapter 1.

## 6.2 Future work

In the future work, new directions might be included to the proposed approach as follows:

1. Increasing the size of the datasets, and performing more experiments.

2. Introducing a preprocessing step to localize and segment the optic disk, because of the confusion between normal and exudates images.

3. Adding color features to SURF, HOG, and LBP descriptors.

4. Extending the proposed approach to deal with more challenging spot lesions, namely microaneurysms.

# Bibliography

[1] Fong, D., Aiello, L., Gardner, T., King, G., Blankenship, G., Cavallerano, J., Ferris, F. and Klein, R. (2004). Retinopathy in diabetes. *Diabetes care*, 27(suppl 1), pp.84--87.

[2] Wild, S., Roglic, G., Green, A., Sicree, R. and King, H. (2004). Global prevalence of diabetes estimates for the year 2000 and projections for 2030. *Diabetes care*, 27(5), pp.1047--1053.

[3] Yen, G. and Leong, W. (2008). A sorting system for hierarchical grading of diabetic fundus images: a preliminary study. *Information Technology in Biomedicine, IEEE Transactions on*, 12(1), pp.118--130.

[4] Gardner, H. (2012). Challenging the pathophysiologic connection between subdural hematoma, retinal hemorrhage, and shaken baby syndrome. *Western Journal of Emergency Medicine*, 13(6), p.535.

[5] Scienceofamd.org, (2014). *Learn |Science Of AMD.* [online] Available at: `http://www.scienceofamd.org/learn/` [Accessed 4 May. 2014].

[6] Niemeijer, M., van Ginneken, B., Russell, S., Suttorp-Schulten, M. and Abramoff, M. (2007). Automated detection and differentiation of drusen, exudates, and cotton-wool spots in digital color fundus photographs for diabetic retinopathy diagnosis. *Investigative ophthalmology & visual science*, 48(5), pp.2260--2267.

[7] Tortora, G. and Derrickson, B., 2009. *Principles of anatomy and physiology.* 12th ed. Chichester: Wiley.

[8] Smith, S., 2002. *The scientist and engineer's guide to digital signal processing.* 1st ed. [San Diego, Calif.]: California Technical Pub.

[9] Litzinger, Thomas C., and Katia Del Rio-Tsonis. Eye Anatomy. *Encyclopedia of Life Sciences* (2002).

[10] Mukherjee, P., 2013. *Ophthalmic assistant.* 1st ed. New Delhi: *Jaypee Brothers Medical Pub.*

[11] Caprette, C., Lee, M., Shine, R., Mokany, A. and Downhower, J., 2004. The origin of snakes (Serpentes) as seen through eye anatomy. *Biological Journal of the Linnean Society*, 81(4), pp.469--482.

[12] Garhart, C. and Lakshminarayanan, V., 2012. Anatomy of the Eye. *Springer Berlin Heidelberg*, pp.73--83.

[13] Heimann, H., Kellner, U. and Foerster, M., 2006. *Atlas of fundus angiography.* 1st ed. Stuttgart: Thieme.

[14] Saine, P. and Tyler, M., 2002. *Ophthalmic photography.* 1st ed. Boston: Butterworth-Heinemann.

[15] Abràmoff, M., Garvin, M. and Sonka, M., 2010. Retinal imaging and image analysis. *Biomedical Engineering, IEEE Reviews in*, 3, pp.169--208.

[16] Retinalmd.com. 2014. *Ophthalmic Ultrasound |Retinal Consultants*. [online] Available at: `http://www.retinalmd.com/en/services/examination/ophthalmic-ultrasound` [Accessed 20 Apr. 2014].

[17] Broecker, E. and Dunbar, M., 2005. Optical coherence tomography: its clinical use for the diagnosis, pathogenesis, and management of macular conditions. *Optometry-Journal of the American Optometric Association*, 76(2), pp.79--101.

[18] Bucca, B., 2014. *Ophthalmology |School of Medicine |University of Colorado Denver*. [online] Ucdenver.edu. Available at: `http://www.ucdenver.edu/academics/colleges/medicalschool/centers/BarbaraDavis/Clinical/Pages/Ophthalmology.aspx` [Accessed 20 Apr. 2014].

[19] Mookiah, M., Acharya, U., Chua, C., Lim, C., Ng, E. and Laude, A., 2013. Computer-aided diagnosis of diabetic retinopathy: A review. *Computers in biology and medicine*, 43(12), pp.2136--2155.

[20] Alghadyan, A., 2011. Diabetic retinopathy–An update. *Saudi Journal of Ophthalmology*, 25(2), pp.99--111.

[21] Cade, W., 2008. Diabetes-related microvascular and macrovascular diseases in the physical therapy setting. *Physical therapy*, 88(11), pp.1322--1335.

[22] Diabetic Retinopathy |Patient Education |The Retina Group of Washington (RGW). 2014. [online] Rgw.com. Available at: `http://rgw.com/patient-education/eye-diseases/diabetic-retinopathy` [Accessed 21 Apr. 2014].

[23] Diabetic Eye Disease, Facts About [NEI Health Information]. 2014. [online] Nei.nih.gov. Available at: `http://www.nei.nih.gov/health/diabetic/retinopathy.asp` [Accessed 21 Apr. 2014].

[24] Ober, M., Klais, C. and Cunningham Jr, E., 2005. Management options for macular edema. *Review of Ophthalmology October*.

[25] Ciulla, T., Amador, A. and Zinman, B., 2003. Diabetic retinopathy and diabetic macular edema pathophysiology, screening, and novel therapies. *Diabetes care*, 26(9), pp.2653--2664.

[26] García, M., Sánchez, C., López, M., Abásolo, D. and Hornero, R. (2009). Neural network based detection of hard exudates in retinal images. *Computer Methods and programs in biomedicine*, 93(1), pp.9--19.

[27] Deepak, K. and Sivaswamy, J. (2012). Automatic assessment of macular edema from color retinal images. *Medical Imaging, IEEE Transactions on*, 31(3), pp.766--776.

[28] Osareh, A., Mirmehdi, M., Thomas, B. and Markham, R. (2003). Automated identification of diabetic retinal exudates in digital colour images. *British Journal of Ophthalmology*, 87(10), pp.1220--1223.

[29] Giancardo, L., Meriaudeau, F., Karnowski, T., Li, Y., Garg, S., Tobin Jr, K. and Chaum, E. (2012). Exudate-based diabetic macular edema detection in fundus images using publicly available datasets. *Medical image analysis*, 16(1), pp.216--226.

[30] Hijazi, M., Coenen, F. and Zheng, Y. (2011). Retinal image classification for the screening of age-related macular degeneration. *Springer Berlin Heidelberg*, pp.325--338.

[31] Hijazi, M., Jiang, C., Coenen, F. and Zheng, Y. (2011). Image classification for age-related macular degeneration screening using hierarchical image decompositions and graph mining. *Springer Berlin Heidelberg*, pp.65--80.

[32] Akram, M.U.; Mujtaba, S.; Tariq, A., Automated drusen segmentation in fundus images for diagnosing age related macular degeneration, *Electronics, Computer and Computation (ICECCO), 2013 International Conference on* , vol., no., pp.17,20, 7-9 Nov. 2013.

[33] Yuanjie Zheng; Vanderbeek, B.; Daniel, E.; Stambolian, D.; Maguire, M.; Brainard, D.; Gee, J., An automated drusen detection system for classifying age-related macular degeneration with color fundus photographs, *Biomedical Imaging (ISBI), 2013 IEEE 10th International Symposium on* , vol., no., pp.1448,1451, 7-11 April 2013.

[34] van Grinsven, M.J.J.P.; Chakravarty, A.; Sivaswamy, J.; Theelen, T.; van Ginneken, B.; Sanchez, C.I., A Bag of Words approach for discriminating between retinal images containing exudates or drusen, *Biomedical Imaging (ISBI), 2013 IEEE 10th International Symposium on* , vol., no., pp.1444,1447, 7-11 April 2013.

[35] Ujjwal; Deepak, K.S.; Chakravarty, A.; Sivaswamy, J., Visual saliency based bright lesion detection and discrimination in retinal images, *Biomedical Imaging (ISBI), 2013 IEEE 10th International Symposium on* , vol., no., pp.1436,1439, 7-11 April 2013.

[36] Bock, R., Meier, J., Nyúl, L., Hornegger, J. and Michelson, G. (2010). Glaucoma risk index: Automated glaucoma detection from color fundus images. *Medical image analysis*, 14(3), pp.471--481.

[37] M. J. Cree, E. Gamble, and D. J. Cornforth, Colour normalisation to reduce inter-patient and intra-patient variability in microaneurysm detection in colour retinal images, in *Workshop on Digital Image Computing*, 2005, pp. 163-169.

[38] Bay, H., Ess, A., Tuytelaars, T. and Van Gool, L. (2008). Speeded-up robust features (SURF). *Computer vision and image understanding*, 110(3), pp.346--359.

[39] Kim, J. (2012). *Advanced methods, techniques, and applications in modeling and simulation.* 1st ed. Tokyo: Springer Berlin Heidelberg.

[40] Evans, C. (2009). Notes on the opensurf library. *University of Bristol, Tech. Rep. CSTR*-09-001, January.

[41] Dalal, N.; Triggs, B., Histograms of oriented gradients for human detection, *Computer Vision and Pattern Recognition, 2005. CVPR 2005. IEEE Computer Society Conference on* , vol.1, no., pp.886,893 vol. 1, 25-25 June 2005.

[42] Felzenszwalb, P., Girshick, R., McAllester, D. and Ramanan, D. (2010). Object detection with discriminatively trained part-based models. *Pattern Analysis and Machine Intelligence, IEEE Transactions on*, 32(9), pp.1627--1645.

[43] VLFeat: *An Open and Portable Library of Computer Vision Algorithms* (2008) by A. Vedaldi, B. Fulkerson.

[44] Ojala, T.; Pietikainen, M.; Harwood, D., Performance evaluation of texture measures with classification based on Kullback discrimination of distributions, *Pattern Recognition, 1994. Vol. 1 - Conference A: Computer Vision & Image Processing., Proceedings of the 12th IAPR International Conference on* , vol.1, no., pp.582,585 vol.1, 9-13 Oct 1994.

[45] Ojala, T., Pietikainen, M. and Harwood, D. (1996). A comparative study of texture measures with classification based on featured distributions. *Pattern Recognition*, 1, pp.51--59.

[46] Huang, D., Jo, K., Lee, H., Kang, H. and Bevilacqua, V. (2009). Proceedings of the Intelligent computing 5th international conference on Emerging intelligent computing technology and applications. *Springer-Verlag.*

[47] Luo, J. (2012). Soft computing in information communication technology. 1st ed. Berlin: *Springer Berlin Heidelberg.*

[48] Wang, X., Wang, L. and Qiao, Y. (2013). A comparative study of encoding, pooling and normalization methods for action recognition. *Springer Berlin Heidelberg*, pp.572--585.

[49] Chang, C. and Lin, C. (2011). LIBSVM: a library for support vector machines. *ACM Transactions on Intelligent Systems and Technology (TIST)*, 2(3), p.27.

[50] Pires, R.; Jelinek, H.F.; Wainer, J.; Goldenstein, S.; Valle, E.; Rocha, A., Assessing the Need for Referral in Automatic Diabetic Retinopathy Detection, *Biomedical Engineering, IEEE Transactions on* , vol.60, no.12, pp.3391,3398, Dec. 2013

[51] Mohamed Aly and Mario Munich and Pietro Perona. (2011). Using More Visual Words in Bag of Words Large Scale Image Search. *Caltech, USA.*

[52] A. Hoover, V. Kouznetsova and M. Goldbaum, Locating Blood Vessels in Retinal Images by Piecewise Threhsold Probing of a Matched Filter Response, *IEEE Transactions on Medical Imaging* , vol. 19 no. 3, pp. 203--210, March 2000. [online] Available at: `http://www.ces.clemson.edu/~ahoover/stare/` [Accessed 18 May. 2014].

[53] M. Niemeijer, J. J. Staal, M. Ginneken B. v., Loog, and M. D. Abramoff. (2004) DRIVE: digital retinal images for vessel extraction. [online] Available at: `http://www.isi.uu.nl/Research/Databases/DRIVE` [Accessed 18 May. 2014].

[54] Prentasic, P., Loncaric, S., Vatavuk, Z., Bencic, G., Subasic, M., Petkovic, T., Dujmovic, L., Malenica-Ravlic, M., Budimlija, N. and Tadic, R. (2013). Diabetic Retinopathy Image Database (DRiDB): a new database for diabetic retinopathy screening programs research. pp.711--716. [online] Available at: `http://www.fer.unizg.hr/ipg/resources/image_database` [Accessed 18 May. 2014].

[55] Messidor.crihan.fr, (2014). Messidor. [online] Available at: `http://messidor.crihan.fr/index-en.php` [Accessed 18 May. 2014].

[56] Budai, A., Hornegger, J. and Michelson, G. (2009). Multiscale approach for blood vessel segmentation on retinal fundus images. *Investigative Ophtalmology and Visual Science*, 50(5), p.325. [online] Available at:`http://www5.cs.fau.de/research/data/fundus-images/` [Accessed 18 May. 2014].